\pdfoutput=1 

\documentclass[11pt]{article}
\usepackage{emnlp2021}

\usepackage{amsmath,mathtools}
\usepackage{times}
\usepackage{graphicx}
\usepackage{wrapfig}
\usepackage{dsfont}
\usepackage{latexsym}
\usepackage[linesnumbered,algoruled,boxed,noend]{algorithm2e}
\usepackage{amssymb}
\usepackage{amsmath}
\usepackage{url} 
\SetKwInOut{Parameter}{parameter}
\usepackage{enumitem}

\usepackage{mathtools}
\usepackage{cleveref}
\usepackage{multirow}
\usepackage{hhline}
\usepackage[export]{adjustbox}
\usepackage{booktabs,array}
\usepackage{amsmath, bm}
\usepackage{color, colortbl}
\usepackage{xcolor}
\usepackage{nicematrix}

\usepackage{listings}

\usepackage[T2A,T1]{fontenc}
\usepackage[utf8]{inputenc}
\usepackage[russian,english]{babel}
\usepackage{textgreek}
\usepackage{CJKutf8}
\usepackage{dsfont}
\usepackage[tikz]{bclogo}

\usepackage{resizegather}
\SetKwInput{KwInput}{Input}
\SetKwInput{KwOutput}{Output}
\newcolumntype{P}[1]{>{\centering\arraybackslash}p{#1}}
\usepackage{booktabs,makecell,tabularx} 
\setitemize{noitemsep,topsep=0pt,parsep=0pt,partopsep=0pt}
\let\oldnl\nl
\definecolor{Gray}{gray}{0.9}
\definecolor{LightCyan}{rgb}{0.88,1,1}
\newcommand{\nonl}{\renewcommand{\nl}{\let\nl\oldnl}}

\usepackage{pifont}

\definecolor{msftBlue}{RGB}{0,164,239}
\definecolor{msftGreen}{RGB}{127,186,0}
\definecolor{msftYello}{RGB}{255,185,0}
\definecolor{msftBlack}{RGB}{0,0,0}
\newcommand{\finding}[1]{
	\begin{bclogo}[couleur= msftBlack!03,  arrondi=0, logo={}, marge=4, couleurBord=msftBlack!10, sousTitre ={\em \textit{\textbf{#1}}}]{} 
	\end{bclogo}
	\vspace{-0.5em}
}

\newcommand{\vx}{\bm{x}}

\newcommand{\obj}{F}

\newcommand{\numClients}{\ensuremath{M}}

\newcommand{\localStep}{\tau}

\newcommand{\data}{\ensuremath{\mathcal{D}}}
\newcommand{\clientDist}{\ensuremath{\mathcal{P}}}

\newcommand{\activeClients}{\mathcal{S}}
\newcommand{\sgrad}{g}

\newcommand{\localChange}{\Delta}
\newcommand{\modelSize}{d}
\newcommand{\E}{\mathbb{E}}

\newcommand{\lr}{\eta}
\newcommand{\slr}{\lr_{s}}

\newcommand{\fedopt}{\textsc{FedOpt}\xspace}

\def\code#1{\texttt{{\footnotesize#1}}}

\usepackage{epigraph}
\usepackage{lipsum}
\usepackage{etoolbox}
\patchcmd{\epigraph}{\@epitext{#1}}{\itshape\@epitext{#1}}{}{}

\newlength{\Width}%
\newlength{\DepthReference}
\newlength{\HeightReference}
\newcommand{\MyColorBox}[2][red]%
{%
    \settowidth{\Width}{#2}%
    \colorbox{#1}%
    {%
        \raisebox{-\DepthReference}%
        {%
                \parbox[b][\HeightReference+\DepthReference][c]{\Width}{\centering#2}%
        }%
    }%
}
\usepackage{subfigure}

\title{
\vspace*{-0.5in}
{{\small \hfill NAACL 2022 Findings}\\
\vspace*{.25in}}
FedNLP: Benchmarking Federated Learning Methods\\ for Natural Language Processing Tasks}

\author{Bill Yuchen Lin$^1$\thanks{~Bill and Chaoyang contributed equally; Xiang and Salman are equal advisors for this work.},~
Chaoyang He$^1$$^*$, Zihang Zeng$^1$, Hulin Wang$^1$, \\
\textbf{Yufen Huang$^1$,  Christophe Dupuy$^2$, Rahul Gupta$^2$,} \\
\textbf{Mahdi Soltanolkotabi$^1$, Xiang Ren$^1$$^*$, Salman Avestimehr$^1$$^*$}\\
University of Southern California$^1$ \qquad Amazon Alexa AI$^2$  \\
{\small{\texttt{\{yuchen.lin,chaoyang.he,saltanol,xiangren,avestime\}@usc.edu~{gupra}@amazon.com }}} \\

}

\begin{document}

\maketitle

\begin{abstract}


Increasing concerns and regulations about data privacy and sparsity necessitate the study of privacy-preserving, decentralized learning methods for natural language processing (NLP) tasks. 
Federated learning (FL) provides promising approaches for a large number of clients (e.g., personal devices or organizations) to collaboratively learn a shared global model to benefit all clients while allowing users to keep their data locally. 
Despite interest in  studying FL methods for NLP tasks, a systematic comparison and analysis is lacking in the literature. 
Herein, we present the FedNLP\footnote{\url{https://github.com/FedML-AI/FedNLP}}, a benchmarking framework for evaluating federated learning  methods on four common  formulations of NLP tasks: text classification, sequence tagging, question answering, and seq2seq generation. 
We propose a universal interface between Transformer-based language models (e.g., BERT, BART) and FL methods  under various non-IID partitioning strategies. 
Our extensive experiments with FedNLP provide empirical comparisons  between FL methods and help us better understand the inherent challenges of this direction. 
The comprehensive analysis points to intriguing and exciting future research aimed at developing FL methods for NLP tasks.


\end{abstract}

\section{Introduction}
\label{sec:intro}
Fine-tuning large pre-trained language models (LMs) such as BERT~\cite{Devlin2019BERTPO} often leads to state-of-the-art performance in many realistic NLP applications (e.g., text classification, named entity recognition, question answering, summarization, etc.), when large-scale, \textit{centralized} training datasets are available.
However, due to the increasing concerns and regulations about data privacy (\textit{e.g.,} GPDR~\cite{gpdr})
emerging data from realistic users have been much more \textit{fragmented} and \textit{distributed},
forming \textit{decentralized private datasets} of multiple ``data silos'' 
(a \textit{data silo} can be viewed as an individual dataset) --- across different clients (e.g., organizations or personal devices).


\begin{figure}[t]
\centering
{\includegraphics[width=1\linewidth]{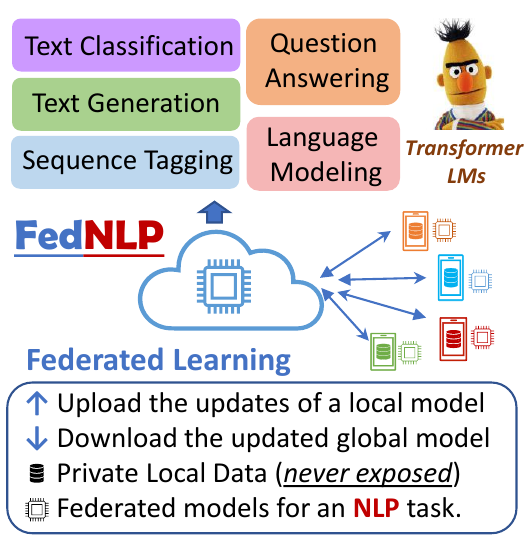}}
\caption{The FedNLP benchmarking framework. }
\label{fig:intro}
\end{figure}

To respect the privacy of the users and abide by these regulations,
we must assume that users' data in a \textit{silo} are not allowed to transfer to a centralized server or other clients.
For example, a client cannot share its private user data (e.g., documents, conversations, questions asked on the website/app) with other clients.
This is a common concern for \textit{organizations} such as hospitals, financial institutions, or legal firms, as well as \textit{personal computing devices} such as smartphones, virtual assistants (e.g., Amazon Alexa, Google Assistant, etc.), or a personal computer. 
However, from a machine learning perspective, models trained on a centralized dataset that combine the data from all organizations or devices usually result in better performance in the NLP domain.
Therefore, it is of vital importance to study NLP problems in such a realistic yet more challenging scenario ---i.e., training data are distributed across different clients and cannot be shared for privacy concerns.

The nascent field of \textit{federated learning}~\cite{Kairouz2019AdvancesAO, Li2020FederatedLC} (FL) aims to enable many individual clients to train their models jointly while keeping their local data \textit{decentralized} and completely \textit{private} from other users or a centralized server. 
A common training schema of FL methods is that each client sends its model parameters to the server, 
which updates and sends back the global model to all clients in each round.
Since the raw data of one client has never been exposed to others, FL is promising as an effective way to address the above challenges, particularly in the NLP domain, where many user-generated text data contain sensitive and/or personal information.

Despite the growing progress in the FL domain, research into and application for NLP has been rather limited. 
There are indeed several recent works on using FL methods for processing medical information extraction tasks~\cite{Sui2020FedEDFL}.
However, such prior work usually has its experimental setup and specific task, making it difficult to fairly compare these FL methods and analyze their performance in other NLP tasks.
We argue that future research in this promising direction (FL for NLP) would highly benefit from a universal benchmarking platform for systematically comparing different FL methods for NLP.
To the best of our knowledge, such a benchmarking platform is still absent from the literature.

Therefore, our goal in this paper is to provide comprehensive comparisons between popular FL methods (e.g., FedAvg~\cite{mcmahan2017communication}, FedOPT~\cite{reddi2020adaptive}, FedProx~\cite{li2018federated}) for four mainstream formulations of NLP tasks: text classification, sequence tagging, question answering, and seq2seq generation.
Although there are few available realistic FL datasets for NLP due to privacy concerns, 
we manage to use existing NLP datasets to create various non-IID data partitions over clients.
These non-IID partitions simulate various kinds of distribution shifts (e.g., label, features, quantities, etc.) over the clients, which often happen in real-world NLP applications.
As for the base NLP models, we use the Transformer architecture~\cite{vaswani2017attentionia} as the backbone and support a wide range of pre-trained LMs such as DistilBERT~\cite{Sanh2019DistilBERTAD},
BERT~\cite{Devlin2019BERTPO}, BART~\cite{lewis2019bart}, etc. 
To conduct extensive experiments, 
we need to support the experiments with multiple options on dimensions such as (1) \textit{task formulations}, (2) \textit{NLP models}, (3) \textit{FL algorithms}, and (4) \textit{non-IID partitions}.
Therefore, we propose FedNLP, a modular framework with universal interfaces among the above four components, which is thus more extensible for supporting future research in FL for NLP.

We aim to unblock the research of FL for NLP with the following two-fold contributions: 
\begin{itemize}
    
    \item \textbf{Evaluation and analysis.} ~ We systematically compare popular federated learning algorithms for mainstream NLP task formulations under multiple non-IID data partitions, which thus provides the first comprehensive understanding. 
    Our analysis reveals that there is a considerably large gap between centralized and decentralized training in various settings. We also analyze the efficiency of different FL methods and model sizes. With our analysis, we highlight several directions to advance FL for NLP. 
    \item \textbf{Resource.} ~The implementation of our experiments also forms a general open-source framework named FedNLP, which is capable of evaluating, analyzing, and developing FL methods for NLP. We also provide decentralized NLP datasets of various task formulations created by various non-IID partitioning strategies for future research.

\end{itemize}

The remainder of this paper is structured as follows.
We introduce the background knowledge of federated learning and several typical FL algorithms
in \S\ref{sec:background}.
Then, we present the proposed non-IID partitioning strategies to create synthetic datasets for different task formulations
in \S\ref{sec:data}.
Our results,  analysis, and findings
are in \S\ref{sec:exp}.
Finally, we discuss related work 
(\S\ref{sec:relatedwork}) 
and conclusions
(\S\ref{sec:conclusion}).

\section{Federated Learning for NLP}
\label{sec:background}
In this section, we first introduce the background knowledge of federated learning (FL) in the context of NLP tasks.
Then, we  illustrate a unified FL framework that we used to study  typical FL algorithms.
Based on this, we build our research framework, a general pipeline for benchmarking and developing FL methods for NLP.

\subsection{Federated Learning Concepts}
\label{ssec:flbasic}

\textit{Federated learning} (FL) is a machine learning paradigm where multiple entities (clients) collaborate in solving a machine learning problem under the coordination of a central server or service provider. 
Each client's raw data is stored locally and not exchanged or transferred; instead, focused updates intended for immediate aggregation are used to achieve the learning objectives \cite{kairouz2019advances}. 
Therefore, federated learning has been seen as a promising direction to decrease the risk of attack and leakage, reduce the difficulty and cost of data movement, and meet the privacy-related data storage regulations.

In the basic conception of federated learning, we would like to minimize the objective function,
\begin{equation}
\begin{split}
\obj(\vx) &= \E_{i \sim \clientDist}[ \obj_i(\vx)], \\ \text{where} \quad \obj_i(\vx) &= \E_{\xi \sim \data_i}[f_i(\vx, \xi)].
\end{split}
\label{eqn:global_obj}
\end{equation}
$\vx \in \mathbb{R}^\modelSize$ represents the parameter for the global model, $\obj_i: \mathbb{R}^\modelSize \rightarrow \mathbb{R}$ denotes the local objective function at client $i$, and $\clientDist$ denotes a distribution on the collection of clients $\mathcal{I}$.  The local loss functions $f_i(\vx,\xi)$ are often the same across all clients, but the local data distribution $\data_i$ will often vary, capturing data heterogeneity.

\emph{Federated averaging} (FedAvg) \cite{mcmahan2017communication} is a common algorithm to solve (\ref{eqn:global_obj}) by dividing the training process into rounds. At the beginning of the $t$-th round ($t \geq 0$), the server broadcasts the current global model $\vx^{(t)}$ to a \emph{cohort} of participants: a random subset of clients from $\activeClients^{(t)}$ which includes \numClients\ clients in total. Then, each sampled client in the round's cohort performs $\localStep_i$ local SGD updates on its own local dataset and sends the local model changes $\Delta_i^{(t)}=\vx_i^{(t,\localStep_i)}-\vx^{(t)}$ to the server. Finally, the server uses the aggregated $\Delta_i^{(t)}$ to update the global model:
$\vx^{(t+1)} 
    = \vx^{(t)} + \frac{\sum_{i \in \activeClients^{(t)}} p_i \Delta_i^{(t)}}{\sum_{i \in \activeClients^{(t)}} p_i}. \label{eqn:upadte_fedavg}
$
where $p_i$ is the relative weight of client $i$. The above procedure will repeat until the algorithm converges. In the \emph{cross-silo} setting where all clients participate in training on every round (each cohort is the entire population), we have $\activeClients^{(t)}=\{1,2,\dots,\numClients\}$. 
Consequently,
we can learn a global model to benefit all clients while preserving their data privacy.

\subsection{Our Unified Framework for FL}
\label{ssec:flmethods}

\begin{algorithm}[ht]
\small
    \DontPrintSemicolon
    \SetKwInput{Input}{Input}
    \SetAlgoLined
    \LinesNumbered
    \Input{Initial model $\vx^{(0)}$, \textsc{ClientOpt}, \textsc{ServerOpt}}
     \For{$t \in \{0,1,\dots,T-1\}$ }{
      Sample a subset $\activeClients^{(t)}$ of clients\;
      
      \For{{\it \bf client} $i \in \activeClients^{(t)}$ {\it \bf in parallel}}{
        Initialize local model $\vx_i^{(t,0)}=\vx^{(t)}$\;
        \For {$k =0,\dots,\localStep_i-1$}{
            Compute local stochastic gradient $\sgrad_i(\vx_i^{(t,k)})$\;
            Perform local update $\vx_i^{(t,k+1)} = \colorbox{blue!18}{\textsc{ClientOpt}}(\vx_i^{(t,k)}, \sgrad_i(\vx_i^{(t,k)}), \lr, t)$\;
        }
        Compute local model changes $\localChange_i^{(t)} = \vx_i^{(t,\localStep_i)} - \vx_i^{(t,0)}$\;
      }
      Aggregate local changes $\localChange^{(t)} = \sum_{i \in \activeClients^{(t)}} p_i \localChange_i^{(t)} / \sum_{i \in \activeClients^{(t)}} p_i$\;
      Update global model $\vx^{(t+1)} = \colorbox{green!18}{\textsc{ServerOpt}}(\vx^{(t)}, -\localChange^{(t)},\slr,t)$\;
     }
     \caption{\fedopt ~\citep{reddi2020adaptive}): A Generic FedAvg Algorithm}
     \label{algo:generalized_fedavg}
\end{algorithm}

In this work, we propose to use FedOPT \citep{reddi2020adaptive}, a generalized version of FedAvg, to build the \texttt{FedNLP} platform. As the pseudo-code presented in \Cref{algo:generalized_fedavg}, the algorithm is parameterized by two gradient-based optimizers: \textsc{ClientOpt} and \textsc{ServerOpt} with client learning rate $\lr$ and server learning rate $\slr$, respectively. While \textsc{ClientOpt} is used to update the local models, \textsc{ServerOpt} treats the negative of aggregated local changes $-\Delta^{(t)}$ as a pseudo-gradient and applies it to the global model. 
This optimization framework generalizes to many aggregation-based FL algorithms and simplifies the system design. 

To make our research general, we explore different combinations of \textsc{SeverOpt} and \textsc{ClientOpt}. The original FedAvg algorithm implicitly sets \textsc{SeverOpt} and \textsc{ClientOpt} to be SGD, with a fixed server learning rate $\slr$ of $1.0$. FedProx \cite{li2018federated}, tackling statistical heterogeneity by restricting the local model updates to be closer to the initial (global) model, can be easily incorporated into this framework by adding L2 regularization for better stability in training. Moreover, given that AdamW~\cite{loshchilov2017decoupled} is widely used in NLP, we set it for \texttt{ClientOpt} and let the \texttt{ServerOpt} be SGD with momentum to reduce the burden of  tuning.

\subsection{The Proposed FedNLP Framework}
To support our research in this paper and other future work in the area of federated learning for NLP,
we build a general research framework named FedNLP, based on the above universal optimization framework.
We here briefly highlight its unique features and leave the details in the following content and a detailed design is shown in App.~\ref{sec:framework}.
First, FedNLP is the very first framework that connects multiple FL algorithms with Transformer-based models, to our best knowledge.
Also, we implement a flexible suite of interfaces to support different types of NLP tasks and models, as well as different non-IID partitioning strategies (Sec.~\ref{ssec:partition}).
To study security and privacy guarantees, we incorporate state-of-the-art secure aggregation algorithms such as LightSecAgg (see ~\ref{app:lightsecagg}).





\section{Benchmarking Setup with FedNLP}
\label{sec:data}

In this section, we introduce the creation of our benchmark datasets from a set of chosen NLP tasks with different non-IID partition methods. We evaluate various FL methods on these datasets.

\subsection{Task Formulations, Datasets, and Models}
\label{ssec:dataselection}

There are numerous NLP applications, but most of them can be categorized based on four mainstream formulations: text classification (TC), sequence tagging (ST), question answering (QA), and seq2seq generation (SS).
The formal definition of each formulation is detailed in Appendix~\S\ref{ssec:nlptasks}.
To cover all formulations while keeping our experiments in a reasonable scope, we select one representative task for each formulation:
\begin{itemize}[leftmargin=*]

    \item \textbf{Text Classification}: \texttt{20Newsgroup}~\cite{Lang95} is a news classification dataset with annotations for 20
labels. {We showcase our FedNLP with this dataset as it has a larger output space (20 labels) than sentiment-analysis datasets, which is an important factor for the label-distribution shift scenarios. }.
    \item \textbf{Sequence Tagging}: \texttt{OntoNotes}~\cite{Pradhan2013TowardsRL} (5.0) is a corpus where sentences have annotations for the entity spans and types. We use it for the named entity recognition task, which is fundamental to information extraction and other applications.
    \item \textbf{QA}: \texttt{MRQA}~\cite{fisch2019mrqa} is a benchmark consisting of 6 popular datasets\footnote{We only use part of the data to demonstrate and verify our hypothesis; we show the train/test split in brackets.}: \texttt{SQuAD}~\cite{Rajpurkar2016SQuAD10}  (8529/431), \texttt{NewsQA}~\cite{Trischler2017NewsQAAM}  (11877/613), \texttt{TriviaQA}~\cite{Joshi2017TriviaQAAL}  
    (4120/176)
    , \texttt{SearchQA}~\cite{Dunn2017SearchQAAN} 
    (9972/499)
    ,
    \texttt{HotpotQA}~\cite{Yang2018HotpotQAAD}
    , and \texttt{NQ}~\cite{Kwiatkowski2019NaturalQA} (9617/795). 
    
    \item \textbf{Seq2Seq Generation}: \texttt{Gigaword}~\cite{DBLP:conf/naacl/2012aw} is a news corpus with headlines that are often used for testing seq2seq models as a summarization task. Other tasks such as dialogue response generation and machine translation can also be adapted to this format.
\end{itemize}

\begin{table}[t]
\centering
\scalebox{0.85
	}{ 
\begin{tabular}{c||c|c|c|c}
\toprule
Task & Txt.Cls. & Seq.Tag. & QA & Seq2Seq \\
\midrule
Dataset & 20News & Onto. & MRQA & Giga. \\ \midrule
\# Training & 11.3k & 50k & 53.9k & 10k \\
\# Test & 7.5k & 5k & 3k & 2k \\
\# Labels & 20 & 37* & N/A & N/A \\
\midrule
Metrics & Acc. & F-1 & F-1 & ROUGE \\
 \bottomrule
\end{tabular}
}
\caption{
Statistics of the selected datasets for our experiments. *37 is the size of the tag vocabulary.
}
\label{tab:dataset}
\end{table}

\smallskip
We show the basic statistics of the above datasets in Table~\ref{tab:dataset}.
Note that our FedNLP as a research platform supports a much wider range of specific tasks of each formulation, while we only introduce the ones used in our experiments here with typical settings.
Moreover, our contribution is more of a general FL+NLP benchmarking platform instead of particular datasets and partitions. 
 
\paragraph{Base NLP Models.}
\label{ssec:nlpmodels}
Fine-tuning pre-trained LMs has been the \textit{de facto} method for NLP research, so we focus on testing Transformer-based architectures in FedNLP. 
Specifically, we choose to use BART~\cite{lewis2019bart}, a text-to-text Transformer model similar to the T5 model~\cite{t5}, for seq2seq tasks.

\subsection{Non-IID Partitioning Strategies}
\label{ssec:partition}
The existing datasets have been used for centralized training in NLP.
As our focus here is to test decentralized learning methods, we need to distribute the existing datasets to a set of clients.
It is the non-IIDness of the client distribution that makes federated learning a challenging problem. 
Thus, we extend the common practice widely used in prior works to the NLP domain for generating synthetic FL benchmarks~\cite{Li2021FederatedLO}.
We first introduce how we control the \textit{label distribution shift} for TC and ST, then the \textit{quantity distribution shift}, and finally how we model the distribution shift in terms of input features for non-classification NLP tasks (e.g., summarization).

\begin{figure}[t]

{\includegraphics[width=1\linewidth]{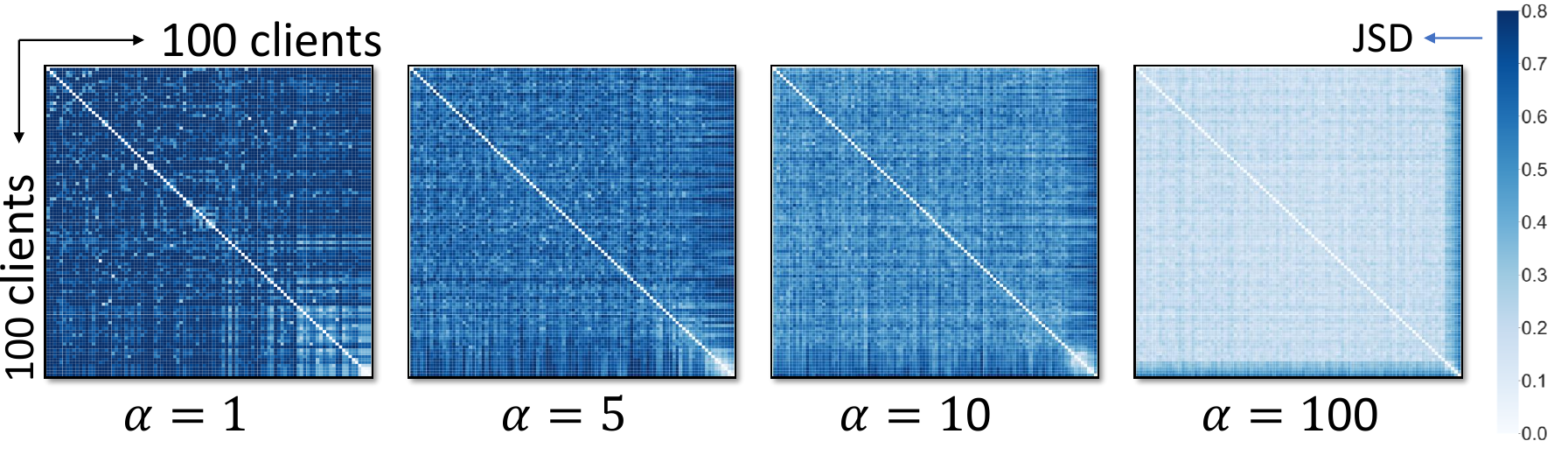}}
\caption{ The \textit{J-S divergence} matrix between 100 clients on the \textit{20News} dataset when $\alpha \in \{1, 5, 10, 100\}$. 
Each sub-figure is a 100x100 symmetric matrix. 
The intensity of a cell $(i, j)$'s color here represents the distance between the label distribution of Client $i$ and $j$. 
It is expected that when $\alpha$ is smaller, the partition over clients is more non-IID in terms of their label distributions. \vspace{-1em}}
\label{fig:labelshiftjsd}
\end{figure}
\begin{figure}[!t]
	\centering
	{\includegraphics[width=0.95\linewidth]{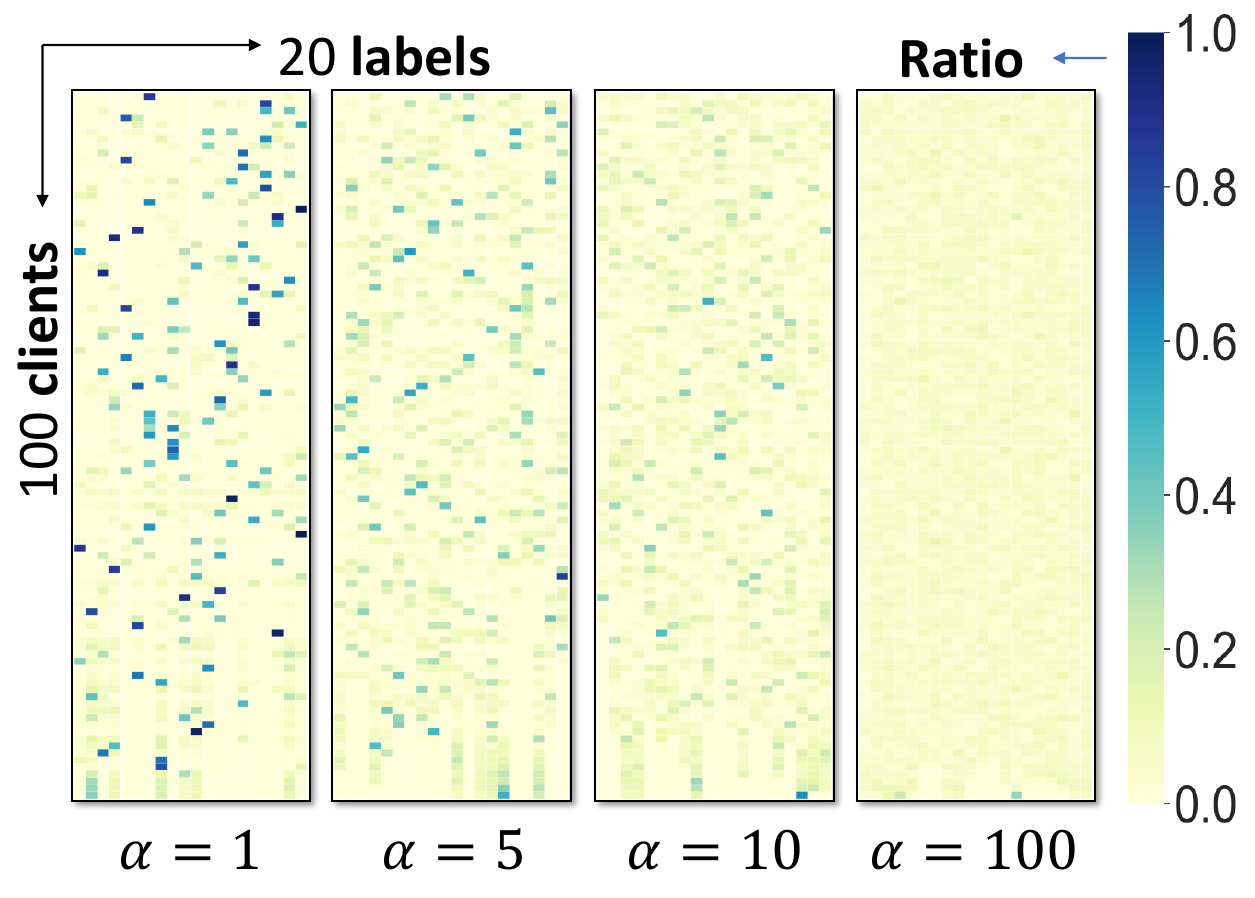}}
	\caption{ Visualizing the \textbf{non-IID label distributions} on \textit{20News} with $\alpha$ being $\{1, 5, 10, 100\}$. Each sub-figure is a 100x20 matrix, where 100 is the number of clients, and 20 is the number of labels. The intensity of a cell here represents the ratio of a particular label in the local data of a client. 
		When $\alpha$ is smaller (1, 5, 10), each client has a relatively unique label distribution, thus the differences between clients are larger; when $\alpha=100$, every client has a nearly uniform label distribution. 
		}
	\label{fig:labeldist}
\end{figure}

\paragraph{Non-IID Label Distributions.}
Here we present how we synthesize the data partitions such that clients share the same (or very similar) number of examples, but have different \textit{label distributions} from each other.
We assume that on every client training, examples are drawn independently with labels following a categorical distribution over $L$ classes parameterized by a vector $\boldsymbol{q}$ $\left(q_{i} \geq 0, i \in[1, L]\right.$ and $\left.\|\boldsymbol{q}\|_{1}=1\right)$. To synthesize a population of non-identical clients, we draw $\boldsymbol{q} \sim \operatorname{Dir}_L(\alpha \boldsymbol{p})$ from a \textit{Dirichlet} distribution, where $\boldsymbol{p}$ characterizes a prior class distribution over $L$ classes, and $\alpha>0$ is a concentration parameter controlling the identicalness among clients.
For each client $C_j$, we draw a $\boldsymbol{q}_j$ as its label distribution and then sample examples without replacement from the global dataset according to $\boldsymbol{q}_j$.
With $\alpha \rightarrow \infty$, all clients have identical distributions to the prior (i.e., uniform distribution); 
with $\alpha \rightarrow 0$, on the other extreme, each client holds examples from only one class chosen at random.
In Fig.~\ref{fig:labelshiftjsd}, we show heatmaps for visualizing the distribution differences between each client.
Figure~\ref{fig:labeldist} shows an example of the concrete label distributions for all clients with different $\alpha$.
We can see that when $\alpha$ is smaller, the overall label distribution shift becomes larger.

\paragraph{Controlling non-IID Quantity.}
It is also common that different clients have very different data quantities while sharing similar label distribution. 
We thus also provide a quantity-level Dirichlet allocation $\boldsymbol{z} \sim \operatorname{Dir}_N(\beta)$ where $N$ is the number of clients.
Then, we can allocate examples in a global dataset to all clients according to the distribution $\boldsymbol{z}$ --- i.e., $|\mathcal{D}_i|= z_i|\mathcal{D}_G|$. 
If we would like to model both quantity and label distribution shift, it is also easy to combine both factors.
Note that one could assume it is a uniform distribution  $\boldsymbol{z}\sim U(N)$, (or $\beta \rightarrow \infty $) if we expect all clients to share a similar number of examples.
A concrete example is shown in Figure~\ref{fig:quantityshift} (Appendix).

\paragraph{Controlling non-IID Features. } 
Although straightforward and effective, the above label-based Dirichlet allocation method has a major limitation --- it is only suitable for text classification tasks where the outputs can be modeled as category-based random variables.
To create synthetic partitions for other non-classification NLP tasks and model distribution shifts,
we thus propose a partition method based on feature clustering.
Specifically,
we use SentenceBERT~\cite{Reimers2019SentenceBERTSE} to encode each example to a dense vector by their text then we apply K-Means clustering to get the cluster label of each example; finally, we use these cluster labels (as if they were classification tasks) to follow the steps in modeling \textit{label distribution shift}. 
There are two obvious benefits of this clustering-based Dirichlet partition method: 
1) It enables us to easily synthesize the FL datasets for non-classification tasks (i.e., ST, QA, SS) as they do not have discrete labels as output space; 
2) The BERT-based clustering results naturally imply different sub-topics of a dataset, and thus feature shift can be seen as a shift of  latent labels --- we can reuse the same method for the label-based Dirichlet partition method.

\begin{table*}[th!]
\centering
\scalebox{0.85
	}{
\begin{tabular}{cc|cc|ccc|c}
\toprule
\textbf{Task} & \textbf{Dataset} & \textbf{Partition} & \textbf{Clients} & \texttt{FedAvg} & \texttt{FedProx} & \texttt{FedOPT} & \textbf{\# Rounds} \\ \midrule
Text Classification & 20news & $\alpha=$1 (label shift) & 100 & 0.5142 & 0.5143 & {0.5349} & 22 \\
Sequence Tagging & OntoNotes & $\alpha=$0.1 (label shift) & 30 & 0.7382 & 0.6731 & 0.7918 & 17 \\
Question Answering & MRQA & natural factor & 6 & 0.2707 & 0.2706 & 0.3280 & 13 \\
Seq2Seq Generation & Gigaword & $\alpha=$0.1 (feature shift) & 100 & 0.3192 & 0.3169 & 0.3037 & 13 \\
\bottomrule
\end{tabular}
}
\caption{The comparisons between different FL methods under the same setting on different NLP tasks. The number of workers per round are 10, expect for the MRQA task, which uses 6. }
\label{tab:results}
\end{table*}

\begin{figure*}[t]
\centering
{\includegraphics[width=1\linewidth]{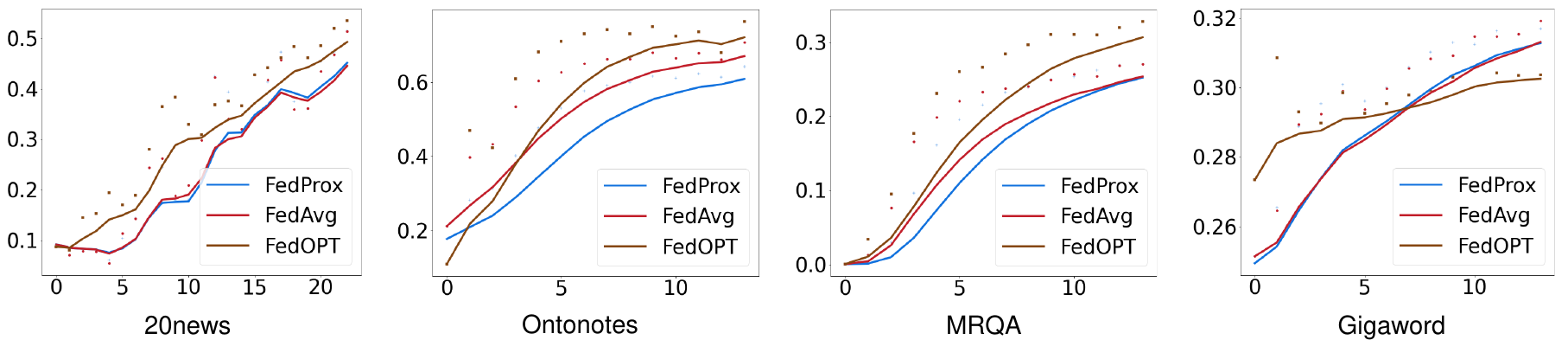}}
\caption{The learning curves of the three FL Methods on four different task formulations. The metrics used for these tasks are accuracy, span-F1, token-F1, and ROUGE respectively; The x-axis is the number of rounds.}
\label{fig:maincurve}
\end{figure*}


\paragraph{Natural Factors} 
For datasets like MRQA, 
we consider a cross-silo setting where each client is associated with a particular sub-dataset (out of the six datasets of the same format), forming a natural distribution shift based on the inherent factors such as data source and annotating style. 

\section{Experimental Results and Analysis}
\label{sec:exp}

In this section, we aim to analyze typical federated learning methods (introduced in our benchmark datasets with multiple dimensions with the base NLP models listed previously.
We put more implementation details and additional results in Appendix. 
We organize our extensive experimental results and findings from the analysis as a collection of research questions with answers. 

\paragraph{Experimental Setup and Hyper-parameters.} 
We use DistilBERT and BART-base for most of our experiments, 
as the former is a distilled version of the BERT model and has a 7x speed improvement over BERT-base on mobile devices --- a common scenario for FL applications; the BART-base model is the most suitable option considering the trade-off between performance and computation cost. 
We leave our implementation details and the selected hyper-parameters in the submitted supplementary materials.

Our experiments cover both cross-device and cross-silo settings. As shown in Table \ref{tab:results}, in the cross-device setting, we use uniform sampling to select 10 clients for each round when the client number in a dataset is very large (e.g., 100). For the cross-silo setting, each round will select the same number of clients (we use 6 for the QA task). The local epoch number is set to 1 for all experiments. 
To make our results reproducible, we use \textit{wandb.ai} to store all experiment logs and hyper-parameters as well as running scripts.

\finding{Q1: How do popular FL methods perform differently under the same setting?}
\smallskip

We compare the three typical FL methods under the same setting (i.e., data partition, communication rounds, etc.) for each task formulation. 
As shown in Table~\ref{tab:results},
we report the results of FedAvg, FedProx, and FedOPT.
We can see that overall FedOPT performs better than the other two methods, with the only exception being in the seq2seq generation task. 
FedAvg and FedProx perform similarly with marginal differences, but FedAvg outperforms FedProx in sequence tagging. 
These two exceptions are surprising findings, as many prior works in the FL community show that FedOPT is generally better than FedProx and FedAvg on vision tasks and datasets.

We conjecture that such inconsistent performance across tasks suggests the difference in terms of the loss functions has a great impact on FL performance.
Seq2seq and sequence tagging tasks usually have more complex loss landscapes than text classification, as they are both typical structured prediction tasks, while the text classification has a much smaller output space.
From Fig.~\ref{fig:maincurve}, 
we see that the FedOPT outperforms the other two methods at the beginning while gradually becoming worse over time.

This tells us that the use of AdamW as the client optimizer may not always be a good choice, especially for a complex task such as the Seq2Seq ones, as its adaptive method for scheduling learning rates might cause implicit conflicts. These observations suggest that federated optimization algorithms need to be tailored for various NLP tasks, and exploring FL-friendly model architecture or loss function can also be promising directions to address these challenges.



\begin{figure}[t]
\centering
{\includegraphics[width=0.95\linewidth]{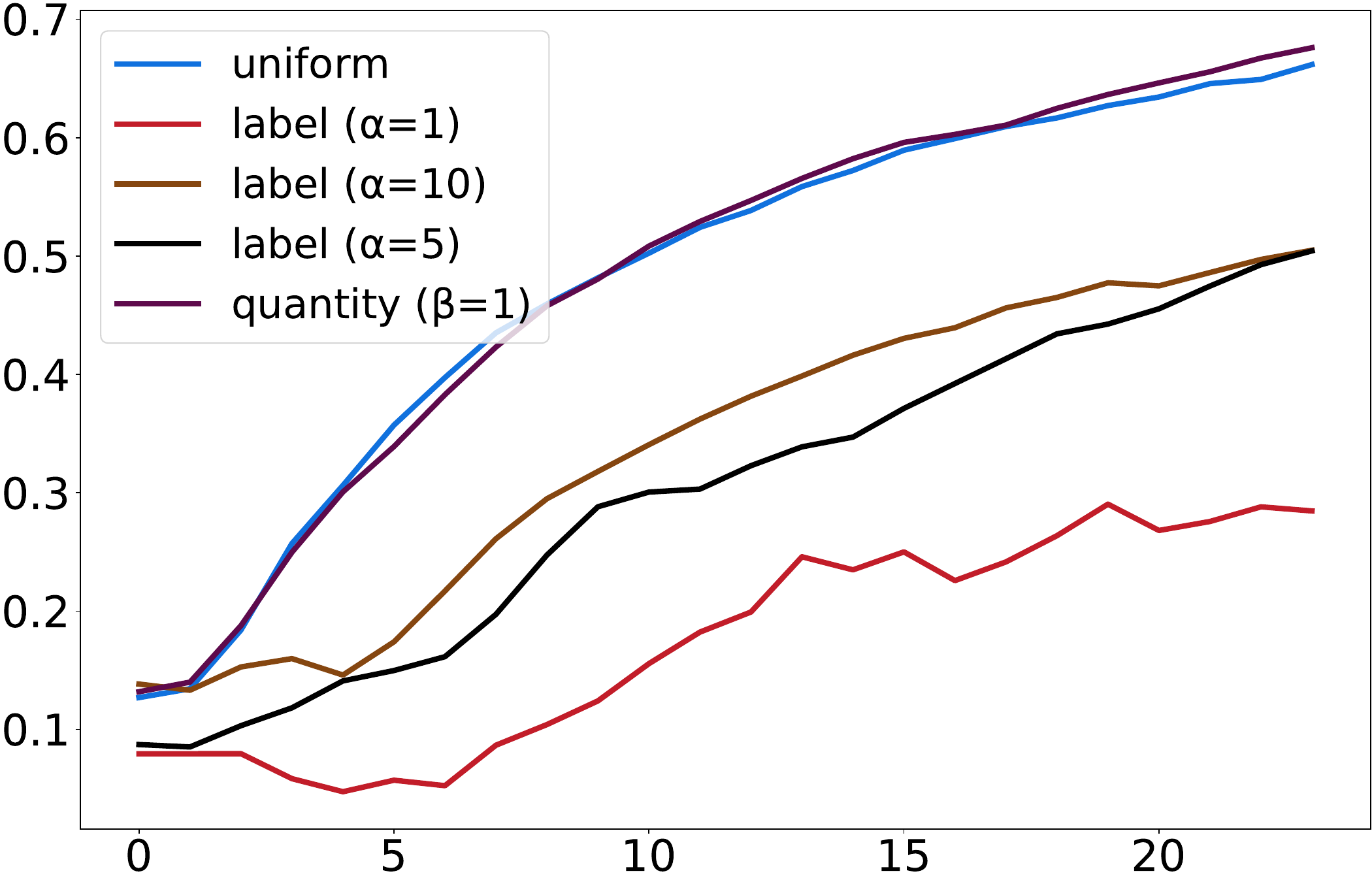}}
\caption{Testing FedOPT with \texttt{DistilBERT} for \texttt{20News} under different data partition strategies. }
\label{fig:curvenews}
\end{figure}

\finding{Q2: How do different non-IID partitions of the same data influence FL performance?}

\smallskip

The FedNLP platform supports users to investigate the performance of an FL algorithm with a wide range of data partitioning strategies, as discussed in \S\ref{ssec:partition}.
Here we look at the training curves of the FedOPT on different partitions, as shown in Figure~\ref{fig:curvenews}.
We reveal several findings: 
\begin{itemize}[leftmargin=*]
    \item When $\alpha$ is smaller (i.e., the partition is more non-IID in terms of their label distribution), the performance tends to degrade, based on the three curves ($\alpha=\{1, 5, 10\}$). 
    \item The variance is also larger when the label distribution shift is larger.
    Both uniform and quantity-skew partitions have a smoother curve, while the variance is smaller for a larger $\alpha$ (e.g., 10).
    \item Quantity skew does not introduce a great challenge for federated learning when the label distribution is closer to the uniform one.
\end{itemize}

These findings suggest that it is important to  design algorithms to mitigate data heterogeneity. 
One promising direction is \textit{personalized} FL, which enables each client to learn its personalized model via adapting its local data distribution and system resources \citep{dinh2020personalized,fallah2020personalized,Li2020DittoFA}. 
 




\begin{table}[t]
\centering
\scalebox{0.85
	}{ 
\begin{tabular}{c|c|cc}
\toprule
Frozen Layers  & \# Tunable Paras.   & Cent.  & FedOpt. \\ \midrule
\texttt{None}  & 67.0M & 86.86 & 55.11       \\
$E$ & 43.1M & 86.19 & 54.86       \\
$E+L_{0}$ & 36.0M & 86.54 & 52.91       \\
$E+L_{0\rightarrow1}$ & 29.0M & 86.52 & 53.92       \\
$E+L_{0\rightarrow2}$ & 21.9M & 85.71 & 52.01       \\
$E+L_{0\rightarrow3}$ & 14.8M & 85.47 & \textit{\underline{30.68}}       \\
$E+L_{0\rightarrow4}$ & 7.7M & 82.76 & \underline{16.63}       \\
$E+L_{0\rightarrow5}$ & 0.6M & \textit{\underline{63.83}} & \underline{12.97 }     \\
\bottomrule
\end{tabular}
}
\caption{
Performance (Acc.\%) on 20news (TC) when different parts of \texttt{DistilBERT} are frozen for centralized training and FedOpt (at 28-th round). $E$ stands for the embedding layer and $L_i$ means the $i$-th layer. The significant lower accuracy are \underline{underlined}.
}
\label{tab:frozen}
\end{table}


\finding{Q3: How does freezing of Transformers influence the FL performance?}

\smallskip
Communication cost is a major concern in the federated learning process.
It is thus natural to consider freezing some Transformer layers of the client models to reduce the size of the trainable parameters that will be transmitted between servers and clients.
To study the influence of freezing layers on the FL performance,
we conduct a series of experiments that freeze the layers from the embedding layer ($E$) to the top layer ($L_5$) of DistilBERT with both centralized training and FedOPT on the text classification task. 
 


\begin{figure}[t]
\centering
{\includegraphics[width=0.95\linewidth]{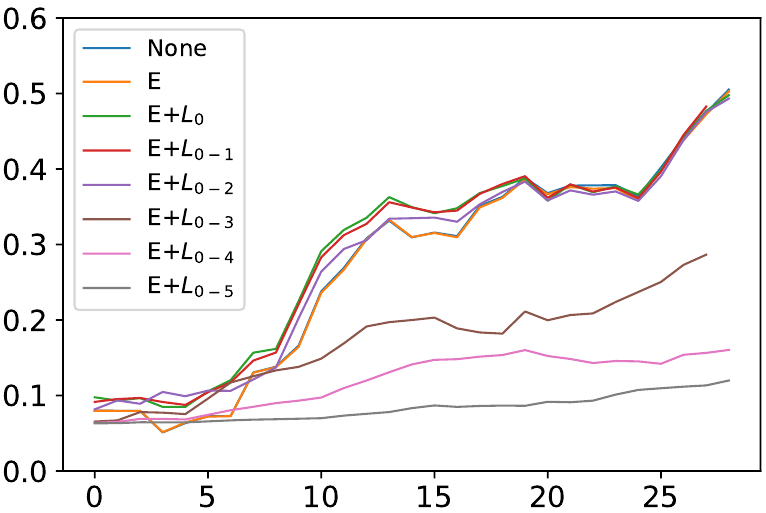}}
\caption{Testing \texttt{FedOPT} with \texttt{DistilBERT} for \texttt{20News} under different frozen layers. }
\label{fig:freeze}
\end{figure}

\begin{figure}[t]
\centering
{\includegraphics[width=1\linewidth]{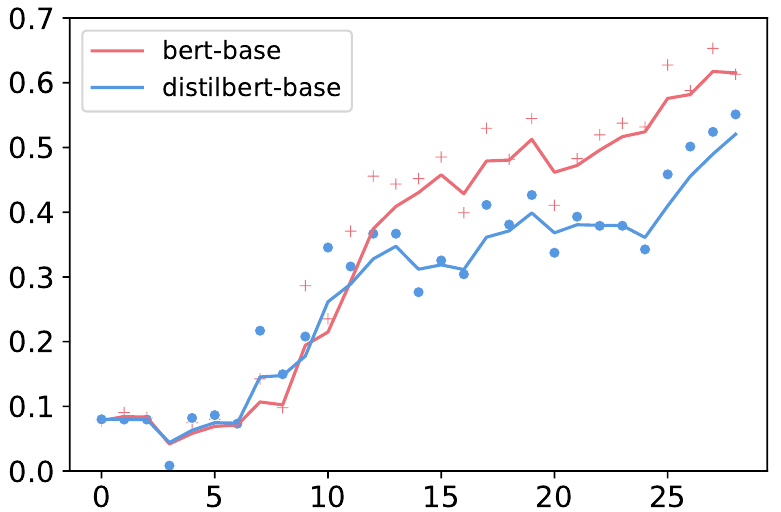}}
\caption{\texttt{FedOPT} for \texttt{20News} with different LMs. }
\label{fig:modelsize}
\end{figure}
We report our results in Table~\ref{tab:frozen} and Figure~\ref{fig:freeze}.
We find that in centralized training, the largest performance gain happens when we unfreeze the last layer, while in FedOPT we have to unfreeze the last three layers to enjoy a comparable performance with the full model. 
This suggests that reducing communication costs via freezing some layers of Transformer LMs is feasible, though one should be aware that the experience in centralized training may not generalize to the FL experiments.


\finding{Q4: Are compact model DistilBERT adequate for FL+NLP?}
\smallskip
We know that BERT has a better performance than DistilBERT for its larger model size. 
However, is it cost-effective to use BERT rather than DistilBERT? 
To study this, 
we compare the performance of both models with FedOPT on text classification, sharing the same setting as the above experiments.
As shown in Figure~\ref{fig:modelsize},
although BERT-base achieves better performance, the performance of DistilBERT is not significantly worse.
Considering the communication cost (BERT-base is almost 2x larger), 
we argue that using DistilBERT is a more cost-effective choice for both experimental analysis and realistic applications.



\section{Related Work}
\label{sec:relatedwork}


\paragraph{FL benchmarks and platforms.} In the last few years a proliferation of frameworks and benchmark datasets have been developed to enable researchers to better explore and study algorithms and modeling for federated learning, both from academia: \text{LEAF}\citep{caldas2018leaf}, \text{FedML} \citep{he2020fedml}, \text{Flower} \citep{beutel2020flower},
and from the industry: \text{PySyft} \citep{ryffel2018generic}, \text{TensorFlow-Federated} (TFF) \citep{TFF2019}, \text{FATE} \citep{yang2019federated}, \text{Clara} \citep{ClaraTraining}, \text{PaddleFL} \citep{ma2019paddlepaddle}, \text{Open FL} \citep{OpenFLFramework}.
However, most  platforms only focus on designing a unified framework for federated learning methods and do not provide a dedicated environment for studying NLP problems with FL methods.
\text{LEAF} \citep{caldas2018leaf} contains a few text datasets, however, it is limited to classification and next-word prediction datasets and does not consider the pre-trained language models. 
We want to provide a dedicated platform for studying FL methods in realistic NLP applications with state-of-the-art language models.

\paragraph{Federated learning in NLP applications.} There are a few prior works that have begun to apply FL methods in privacy-oriented NLP applications.
For example, 
federated learning has been applied to many keyboard-related applications including \cite{Hard2018FederatedLF,Stremmel2020PretrainingFT,Leroy2019FederatedLF,Ramaswamy2019FederatedLF,Yang2018APPLIEDFL}, sentence-level text intent classification using Text-CNN 
\cite{Zhu2020EmpiricalSO}, and pretraining and fine-tuning of BERT using medical data from
multiple silos without fetching all data to the same place \cite{Liu2020FederatedPA}. 
FL methods also have been proposed to train high-quality language models that can outperform the models trained without federated learning  \cite{Ji2019LearningPN,Chen2019FederatedLO}. 
Besides these applications, some work has been done in  medical relation extractions \cite{Ge2020FedNERPM} and medical name entity recognition  \cite{Sui2020FedEDFL}.
These methods use federated learning to preserve the privacy of sensitive medical data and learn data on different platforms, excluding the need for exchanging data between different platforms.

Our work aims to provide a unified platform for studying various NLP applications in a shared environment so that researchers can better design new FL methods either for a specific NLP task or as a general-purpose model.
The aforementioned prior works would thus be a particular instance of the settings supported by the FedNLP platform.

\section{Conclusion and Future Directions}
\label{sec:conclusion}
Our key contribution is providing
a thorough and insightful empirical analysis of existing federated learning algorithms in the context of NLP models.
Notably, We compare typical FL methods for four NLP task formulations under multiple non-IID data partitions. 
Our findings reveal both promise and the challenges of FL for NLP. 
In addition, we also provide
a suite of resources to support future research in FL for NLP
(e.g., a unifying framework for connecting Transformer models with popular FL methods and different non-IID partition
strategies). 
Thus, we believe our well-maintained open-source codebase to support future work in this area.

Promising future directions in FL for NLP include: 1) minimizing the performance gap, 2) improving the system efficiency and scalability, 3) trustworthy and privacy-preserving NLP, 4) personalized FL methods for NLP, etc. (Please see Appendix~\ref{app:future_directions} for more details.)




\section*{Ethical Considerations and Limitations(*)}

\paragraph{Ethical considerations.}
The key motivation of FedNLP (and FL) is to protect the data privacy of general users by keeping their data on their own devices while benefiting from a shared model from a broader community. 
Among the risks that need to be considered in any deployment of NLP  are that responses may be wrong, or biased, in ways that would lead to improperly justified decisions. 
Although in our view the current technology is still relatively immature, and unlikely to be fielded in applications that would cause harm of this sort, 
it is desirable that FedNLP methods provide audit trails, 
and recourse so that their predictions can be explained to and critiqued by affected parties.  

\paragraph{Limitations.}
One limitation of our work is that we have not analyzed the privacy leakage of  FL methods.
We argue that novel privacy-centric measures are orthogonal to the development of FL methods, which is beyond the scope of our work. 
How to fairly analyze the privacy leakage is now still an open problem for both FL and NLP, and it is only possible to study this when we have an existing platform like FedNLP.

\section*{Acknowledgements}
This work is supported in part by a research grant and an Amazon ML Fellowship from USC-Amazon Center on Secure and Trustworthy AI (\url{https://trustedai.usc.edu}).
Xiang Ren is supported in part by the Office of the Director of National Intelligence (ODNI), Intelligence Advanced Research Projects Activity (IARPA), via Contract No. 2019-19051600007, the DARPA MCS program under Contract No. N660011924033, the Defense Advanced Research Projects Agency with award W911NF-19-20271, NSF IIS 2048211, NSF SMA 1829268, and gift awards from Google, Amazon, JP Morgan and Sony.   
Mahdi Soltanolkotabi is supported by the Packard Fellowship in Science and Engineering, 
a Sloan Research Fellowship in Mathematics, an NSF-CAREER under award \#1846369, DARPA Learning with Less Labels (LwLL) and FastNICS programs, and NSF-CIF awards \#1813877 and \#2008443.

{{\bibliography{aaai22_rebiber} }}

\begin{thebibliography}{69}
\expandafter\ifx\csname natexlab\endcsname\relax\def\natexlab#1{#1}\fi

\bibitem[{DBL(2012)}]{DBLP:conf/naacl/2012aw}
 2012.
\newblock \emph{Proceedings of the Joint Workshop on Automatic Knowledge Base
  Construction and Web-scale Knowledge Extraction, AKBC-WEKEX@NAACL-HLT 2012,
  Montr{\`{e}}al, Canada, June 7-8, 2012}.

\bibitem[{Bell et~al.(2020)Bell, Bonawitz, Gasc{\'o}n, Lepoint, and
  Raykova}]{bell2020secure}
James~Henry Bell, Kallista~A Bonawitz, Adri{\`a} Gasc{\'o}n, Tancr{\`e}de
  Lepoint, and Mariana Raykova. 2020.
\newblock Secure single-server aggregation with (poly) logarithmic overhead.
\newblock In \emph{Proceedings of the 2020 ACM SIGSAC Conference on Computer
  and Communications Security}.

\bibitem[{Beutel et~al.(2020)Beutel, Topal, Mathur, Qiu, Parcollet, and
  Lane}]{beutel2020flower}
Daniel~J Beutel, Taner Topal, Akhil Mathur, Xinchi Qiu, Titouan Parcollet, and
  Nicholas~D Lane. 2020.
\newblock Flower: A friendly federated learning research framework.
\newblock \emph{ArXiv preprint}.

\bibitem[{Bonawitz et~al.(2017)Bonawitz, Ivanov, Kreuter, Marcedone, McMahan,
  Patel, Ramage, Segal, and Seth}]{bonawitz2017practical}
Keith Bonawitz, Vladimir Ivanov, Ben Kreuter, Antonio Marcedone, H~Brendan
  McMahan, Sarvar Patel, Daniel Ramage, Aaron Segal, and Karn Seth. 2017.
\newblock Practical secure aggregation for privacy-preserving machine learning.
\newblock In \emph{proceedings of the 2017 ACM SIGSAC Conference on Computer
  and Communications Security}.

\bibitem[{Caldas et~al.(2018)Caldas, Wu, Li, Kone{\v{c}}n{\`y}, McMahan, Smith,
  and Talwalkar}]{caldas2018leaf}
Sebastian Caldas, Peter Wu, Tian Li, Jakub Kone{\v{c}}n{\`y}, H~Brendan
  McMahan, Virginia Smith, and Ameet Talwalkar. 2018.
\newblock Leaf: A benchmark for federated settings.
\newblock \emph{ArXiv preprint}.

\bibitem[{Chen et~al.(2019)Chen, Suresh, Mathews, Wong, Allauzen, Beaufays, and
  Riley}]{Chen2019FederatedLO}
Mingqing Chen, Ananda~Theertha Suresh, Rajiv Mathews, Adeline Wong, Cyril
  Allauzen, Fran{\c{c}}oise Beaufays, and Michael Riley. 2019.
\newblock \href {https://doi.org/10.18653/v1/K19-1012} {Federated learning of
  n-gram language models}.
\newblock In \emph{Proceedings of the 23rd Conference on Computational Natural
  Language Learning (CoNLL)}, pages 121--130, Hong Kong, China. Association for
  Computational Linguistics.

\bibitem[{Devlin et~al.(2019)Devlin, Chang, Lee, and
  Toutanova}]{Devlin2019BERTPO}
Jacob Devlin, Ming-Wei Chang, Kenton Lee, and Kristina Toutanova. 2019.
\newblock \href {https://doi.org/10.18653/v1/N19-1423} {{BERT}: Pre-training of
  deep bidirectional transformers for language understanding}.
\newblock In \emph{Proceedings of the 2019 Conference of the North {A}merican
  Chapter of the Association for Computational Linguistics: Human Language
  Technologies, Volume 1 (Long and Short Papers)}, pages 4171--4186,
  Minneapolis, Minnesota. Association for Computational Linguistics.

\bibitem[{Dinh et~al.(2020)Dinh, Tran, and Nguyen}]{dinh2020personalized}
Canh~T. Dinh, Nguyen~H. Tran, and Tuan~Dung Nguyen. 2020.
\newblock \href
  {https://proceedings.neurips.cc/paper/2020/hash/f4f1f13c8289ac1b1ee0ff176b56fc60-Abstract.html}
  {Personalized federated learning with moreau envelopes}.
\newblock In \emph{Advances in Neural Information Processing Systems 33: Annual
  Conference on Neural Information Processing Systems 2020, NeurIPS 2020,
  December 6-12, 2020, virtual}.

\bibitem[{Dunn et~al.(2017)Dunn, Sagun, Higgins, G{\"u}ney, Cirik, and
  Cho}]{Dunn2017SearchQAAN}
Matthew Dunn, Levent Sagun, Mike Higgins, V.~U. G{\"u}ney, Volkan Cirik, and
  Kyunghyun Cho. 2017.
\newblock Searchqa: A new q\&a dataset augmented with context from a search
  engine.
\newblock \emph{ArXiv}.

\bibitem[{Elkordy and Avestimehr(2020)}]{Elkordy2020SecureAW}
A.~Elkordy and A.~Avestimehr. 2020.
\newblock Secure aggregation with heterogeneous quantization in federated
  learning.
\newblock \emph{ArXiv}.

\bibitem[{et~al(2019)}]{Kairouz2019AdvancesAO}
P.~Kairouz et~al. 2019.
\newblock Advances and open problems in federated learning.
\newblock \emph{ArXiv}.

\bibitem[{Fallah et~al.(2020)Fallah, Mokhtari, and
  Ozdaglar}]{fallah2020personalized}
Alireza Fallah, Aryan Mokhtari, and Asuman Ozdaglar. 2020.
\newblock Personalized federated learning: A meta-learning approach.
\newblock \emph{ArXiv preprint}.

\bibitem[{Fisch et~al.(2019)Fisch, Talmor, Jia, Seo, Choi, and
  Chen}]{fisch2019mrqa}
Adam Fisch, Alon Talmor, Robin Jia, Minjoon Seo, Eunsol Choi, and Danqi Chen.
  2019.
\newblock \href {https://doi.org/10.18653/v1/D19-5801} {{MRQA} 2019 shared
  task: Evaluating generalization in reading comprehension}.
\newblock In \emph{Proceedings of the 2nd Workshop on Machine Reading for
  Question Answering}, pages 1--13, Hong Kong, China. Association for
  Computational Linguistics.

\bibitem[{Ge et~al.(2020)Ge, Wu, Wu, Qi, Huang, and Xie}]{Ge2020FedNERPM}
Suyu Ge, Fangzhao Wu, Chuhan Wu, Tao Qi, Yongfeng Huang, and X.~Xie. 2020.
\newblock Fedner: Privacy-preserving medical named entity recognition with
  federated learning.
\newblock \emph{ArXiv}.

\bibitem[{Hard et~al.(2018)Hard, Rao, Mathews, Beaufays, Augenstein, Eichner,
  Kiddon, and Ramage}]{Hard2018FederatedLF}
Andrew Hard, K.~Rao, Rajiv Mathews, F.~Beaufays, S.~Augenstein, Hubert Eichner,
  Chlo{\'e} Kiddon, and D.~Ramage. 2018.
\newblock Federated learning for mobile keyboard prediction.
\newblock \emph{ArXiv}.

\bibitem[{He et~al.(2020{\natexlab{a}})He, Annavaram, and
  Avestimehr}]{he2020fednas}
Chaoyang He, Murali Annavaram, and Salman Avestimehr. 2020{\natexlab{a}}.
\newblock Fednas: Federated deep learning via neural architecture search.

\bibitem[{He et~al.(2020{\natexlab{b}})He, Annavaram, and
  Avestimehr}]{he2020group}
Chaoyang He, Murali Annavaram, and Salman Avestimehr. 2020{\natexlab{b}}.
\newblock \href
  {https://proceedings.neurips.cc/paper/2020/hash/a1d4c20b182ad7137ab3606f0e3fc8a4-Abstract.html}
  {Group knowledge transfer: Federated learning of large cnns at the edge}.
\newblock In \emph{Advances in Neural Information Processing Systems 33: Annual
  Conference on Neural Information Processing Systems 2020, NeurIPS 2020,
  December 6-12, 2020, virtual}.

\bibitem[{He et~al.(2021)He, Balasubramanian, Ceyani, Rong, Zhao, Huang,
  Annavaram, and Avestimehr}]{He2021FedGraphNNAF}
Chaoyang He, Keshav Balasubramanian, Emir Ceyani, Yu~Rong, Peilin Zhao, Junzhou
  Huang, M.~Annavaram, and S.~Avestimehr. 2021.
\newblock Fedgraphnn: A federated learning system and benchmark for graph
  neural networks.

\bibitem[{He et~al.(2020{\natexlab{c}})He, Li, So, Zeng, Zhang, Wang, Wang,
  Vepakomma, Singh, Qiu, Zhu, Wang, Shen, Zhao, Kang, Liu, Raskar, Yang,
  Annavaram, and Avestimehr}]{he2020fedml}
Chaoyang He, Songze Li, Jinhyun So, Xiao Zeng, Mi~Zhang, Hongyi Wang, Xiaoyang
  Wang, Praneeth Vepakomma, Abhishek Singh, Hang Qiu, Xinghua Zhu, Jianzong
  Wang, Li~Shen, Peilin Zhao, Yan Kang, Yang Liu, Ramesh Raskar, Qiang Yang,
  Murali Annavaram, and Salman Avestimehr. 2020{\natexlab{c}}.
\newblock Fedml: A research library and benchmark for federated machine
  learning.
\newblock \emph{ArXiv preprint}.

\bibitem[{He et~al.(2019)He, Tan, Tang, Qiu, and Liu}]{he2019central}
Chaoyang He, Conghui Tan, Hanlin Tang, Shuang Qiu, and Ji~Liu. 2019.
\newblock Central server free federated learning over single-sided trust social
  networks.
\newblock \emph{ArXiv preprint}.

\bibitem[{He et~al.(2020{\natexlab{d}})He, Ye, Shen, and Zhang}]{MiLeNAS2020}
Chaoyang He, Haishan Ye, Li~Shen, and Tong Zhang. 2020{\natexlab{d}}.
\newblock \href {https://doi.org/10.1109/CVPR42600.2020.01201} {Milenas:
  Efficient neural architecture search via mixed-level reformulation}.
\newblock In \emph{2020 {IEEE/CVF} Conference on Computer Vision and Pattern
  Recognition, {CVPR} 2020, Seattle, WA, USA, June 13-19, 2020}, pages
  11990--11999. {IEEE}.

\bibitem[{Ingerman and Ostrowski(2019)}]{TFF2019}
Alex Ingerman and Krzys Ostrowski. 2019.
\newblock \emph{TensorFlow Federated}.

\bibitem[{Intel®(2021)}]{OpenFLFramework}
Intel®. 2021.
\newblock Intel® open federated learning.

\bibitem[{Ji et~al.(2019)Ji, Pan, Long, Li, Jiang, and
  Huang}]{Ji2019LearningPN}
Shaoxiong Ji, Shirui Pan, Guodong Long, Xue Li, Jing Jiang, and Zi~Huang. 2019.
\newblock Learning private neural language modeling with attentive aggregation.
\newblock \emph{2019 International Joint Conference on Neural Networks
  (IJCNN)}.

\bibitem[{Joshi et~al.(2017)Joshi, Choi, Weld, and
  Zettlemoyer}]{Joshi2017TriviaQAAL}
Mandar Joshi, Eunsol Choi, Daniel Weld, and Luke Zettlemoyer. 2017.
\newblock \href {https://doi.org/10.18653/v1/P17-1147} {{T}rivia{QA}: A large
  scale distantly supervised challenge dataset for reading comprehension}.
\newblock In \emph{Proceedings of the 55th Annual Meeting of the Association
  for Computational Linguistics (Volume 1: Long Papers)}, pages 1601--1611,
  Vancouver, Canada. Association for Computational Linguistics.

\bibitem[{Kairouz et~al.(2019)Kairouz, McMahan, Avent, Bellet, Bennis, Bhagoji,
  Bonawitz, Charles, Cormode, Cummings et~al.}]{kairouz2019advances}
Peter Kairouz, H~Brendan McMahan, Brendan Avent, Aur{\'e}lien Bellet, Mehdi
  Bennis, Arjun~Nitin Bhagoji, Keith Bonawitz, Zachary Charles, Graham Cormode,
  Rachel Cummings, et~al. 2019.
\newblock Advances and open problems in federated learning.
\newblock \emph{ArXiv preprint}.

\bibitem[{Kwiatkowski et~al.(2019)Kwiatkowski, Palomaki, Redfield, Collins,
  Parikh, Alberti, Epstein, Polosukhin, Devlin, Lee, Toutanova, Jones, Kelcey,
  Chang, Dai, Uszkoreit, Le, and Petrov}]{Kwiatkowski2019NaturalQA}
Tom Kwiatkowski, Jennimaria Palomaki, Olivia Redfield, Michael Collins, Ankur
  Parikh, Chris Alberti, Danielle Epstein, Illia Polosukhin, Jacob Devlin,
  Kenton Lee, Kristina Toutanova, Llion Jones, Matthew Kelcey, Ming-Wei Chang,
  Andrew~M. Dai, Jakob Uszkoreit, Quoc Le, and Slav Petrov. 2019.
\newblock \href {https://doi.org/10.1162/tacl_a_00276} {Natural questions: A
  benchmark for question answering research}.
\newblock \emph{Transactions of the Association for Computational Linguistics},
  7:452--466.

\bibitem[{Lang(1995)}]{Lang95}
Ken Lang. 1995.
\newblock Newsweeder: Learning to filter netnews.
\newblock In \emph{Proc. of ICML}.

\bibitem[{Leroy et~al.(2019)Leroy, Coucke, Lavril, Gisselbrecht, and
  Dureau}]{Leroy2019FederatedLF}
David Leroy, Alice Coucke, Thibaut Lavril, Thibault Gisselbrecht, and Joseph
  Dureau. 2019.
\newblock \href {https://doi.org/10.1109/ICASSP.2019.8683546} {Federated
  learning for keyword spotting}.
\newblock In \emph{{IEEE} International Conference on Acoustics, Speech and
  Signal Processing, {ICASSP} 2019, Brighton, United Kingdom, May 12-17, 2019},
  pages 6341--6345. {IEEE}.

\bibitem[{Lewis et~al.(2020)Lewis, Liu, Goyal, Ghazvininejad, Mohamed, Levy,
  Stoyanov, and Zettlemoyer}]{lewis2019bart}
Mike Lewis, Yinhan Liu, Naman Goyal, Marjan Ghazvininejad, Abdelrahman Mohamed,
  Omer Levy, Veselin Stoyanov, and Luke Zettlemoyer. 2020.
\newblock \href {https://doi.org/10.18653/v1/2020.acl-main.703} {{BART}:
  Denoising sequence-to-sequence pre-training for natural language generation,
  translation, and comprehension}.
\newblock In \emph{Proceedings of the 58th Annual Meeting of the Association
  for Computational Linguistics}, pages 7871--7880, Online. Association for
  Computational Linguistics.

\bibitem[{Li et~al.(2021{\natexlab{a}})Li, Diao, Chen, and
  He}]{Li2021FederatedLO}
Q.~Li, Yiqun Diao, Quan Chen, and Bingsheng He. 2021{\natexlab{a}}.
\newblock Federated learning on non-iid data silos: An experimental study.
\newblock \emph{ArXiv}.

\bibitem[{Li et~al.(2021{\natexlab{b}})Li, Hu, Beirami, and
  Smith}]{Li2020DittoFA}
Tian Li, Shengyuan Hu, Ahmad Beirami, and Virginia Smith. 2021{\natexlab{b}}.
\newblock \href {http://proceedings.mlr.press/v139/li21h.html} {Ditto: Fair and
  robust federated learning through personalization}.
\newblock In \emph{Proceedings of the 38th International Conference on Machine
  Learning, {ICML} 2021, 18-24 July 2021, Virtual Event}, volume 139 of
  \emph{Proceedings of Machine Learning Research}, pages 6357--6368. {PMLR}.

\bibitem[{Li et~al.(2020{\natexlab{a}})Li, Sahu, Talwalkar, and
  Smith}]{Li2020FederatedLC}
Tian Li, Anit~Kumar Sahu, Ameet Talwalkar, and V.~Smith. 2020{\natexlab{a}}.
\newblock Federated learning: Challenges, methods, and future directions.
\newblock \emph{IEEE Signal Processing Magazine}.

\bibitem[{Li et~al.(2020{\natexlab{b}})Li, Sahu, Zaheer, Sanjabi, Talwalkar,
  and Smith}]{li2018federated}
Tian Li, Anit~Kumar Sahu, Manzil Zaheer, Maziar Sanjabi, Ameet Talwalkar, and
  Virginia Smith. 2020{\natexlab{b}}.
\newblock \href {https://proceedings.mlsys.org/book/316.pdf} {Federated
  optimization in heterogeneous networks}.
\newblock In \emph{Proceedings of Machine Learning and Systems 2020, MLSys
  2020, Austin, TX, USA, March 2-4, 2020}. mlsys.org.

\bibitem[{Li and Liang(2021)}]{Li2021PrefixTuningOC}
Xiang~Lisa Li and Percy Liang. 2021.
\newblock \href {https://doi.org/10.18653/v1/2021.acl-long.353} {Prefix-tuning:
  Optimizing continuous prompts for generation}.
\newblock In \emph{Proceedings of the 59th Annual Meeting of the Association
  for Computational Linguistics and the 11th International Joint Conference on
  Natural Language Processing (Volume 1: Long Papers)}, pages 4582--4597,
  Online. Association for Computational Linguistics.

\bibitem[{Liu and Miller(2020)}]{Liu2020FederatedPA}
D.~Liu and T.~Miller. 2020.
\newblock Federated pretraining and fine tuning of bert using clinical notes
  from multiple silos.
\newblock \emph{ArXiv}.

\bibitem[{Loshchilov and Hutter(2019)}]{loshchilov2017decoupled}
Ilya Loshchilov and Frank Hutter. 2019.
\newblock \href {https://openreview.net/forum?id=Bkg6RiCqY7} {Decoupled weight
  decay regularization}.
\newblock In \emph{7th International Conference on Learning Representations,
  {ICLR} 2019, New Orleans, LA, USA, May 6-9, 2019}. OpenReview.net.

\bibitem[{Lyu et~al.(2020)Lyu, Yu, Ma, Sun, Zhao, Yang, and
  Yu}]{lyu2020privacy}
Lingjuan Lyu, Han Yu, Xingjun Ma, Lichao Sun, Jun Zhao, Qiang Yang, and
  Philip~S Yu. 2020.
\newblock Privacy and robustness in federated learning: Attacks and defenses.
\newblock \emph{ArXiv preprint}.

\bibitem[{Ma et~al.(2019)Ma, Yu, Wu, and Wang}]{ma2019paddlepaddle}
Yanjun Ma, Dianhai Yu, Tian Wu, and Haifeng Wang. 2019.
\newblock Paddlepaddle: An open-source deep learning platform from industrial
  practice.
\newblock \emph{Frontiers of Data and Domputing}, (1).

\bibitem[{McMahan et~al.(2017{\natexlab{a}})McMahan, Moore, Ramage, Hampson,
  and y~Arcas}]{mcmahan2017communication}
Brendan McMahan, Eider Moore, Daniel Ramage, Seth Hampson, and
  Blaise~Ag{\"{u}}era y~Arcas. 2017{\natexlab{a}}.
\newblock \href {http://proceedings.mlr.press/v54/mcmahan17a.html}
  {Communication-efficient learning of deep networks from decentralized data}.
\newblock In \emph{Proceedings of the 20th International Conference on
  Artificial Intelligence and Statistics, {AISTATS} 2017, 20-22 April 2017,
  Fort Lauderdale, FL, {USA}}, volume~54 of \emph{Proceedings of Machine
  Learning Research}, pages 1273--1282. {PMLR}.

\bibitem[{McMahan et~al.(2017{\natexlab{b}})McMahan, Moore, Ramage, Hampson,
  and y~Arcas}]{McMahan2017CommunicationEfficientLO}
Brendan McMahan, Eider Moore, Daniel Ramage, Seth Hampson, and
  Blaise~Ag{\"{u}}era y~Arcas. 2017{\natexlab{b}}.
\newblock \href {http://proceedings.mlr.press/v54/mcmahan17a.html}
  {Communication-efficient learning of deep networks from decentralized data}.
\newblock In \emph{Proceedings of the 20th International Conference on
  Artificial Intelligence and Statistics, {AISTATS} 2017, 20-22 April 2017,
  Fort Lauderdale, FL, {USA}}, volume~54 of \emph{Proceedings of Machine
  Learning Research}, pages 1273--1282. {PMLR}.

\bibitem[{NVIDIA(2019)}]{ClaraTraining}
NVIDIA. 2019.
\newblock Nvidia clara.

\bibitem[{Pradhan et~al.(2013)Pradhan, Moschitti, Xue, Ng, Bj{\"o}rkelund,
  Uryupina, Zhang, and Zhong}]{Pradhan2013TowardsRL}
Sameer Pradhan, Alessandro Moschitti, Nianwen Xue, Hwee~Tou Ng, Anders
  Bj{\"o}rkelund, Olga Uryupina, Yuchen Zhang, and Zhi Zhong. 2013.
\newblock \href {https://aclanthology.org/W13-3516} {Towards robust linguistic
  analysis using {O}nto{N}otes}.
\newblock In \emph{Proceedings of the Seventeenth Conference on Computational
  Natural Language Learning}, pages 143--152, Sofia, Bulgaria. Association for
  Computational Linguistics.

\bibitem[{Prakash and Avestimehr(2020)}]{prakash2020mitigating}
Saurav Prakash and Amir~Salman Avestimehr. 2020.
\newblock Mitigating byzantine attacks in federated learning.
\newblock \emph{ArXiv preprint}.

\bibitem[{Prakash et~al.(2020)Prakash, Dhakal, Akdeniz, Yona, Talwar,
  Avestimehr, and Himayat}]{prakash2020coded}
Saurav Prakash, Sagar Dhakal, Mustafa~Riza Akdeniz, Yair Yona, Shilpa Talwar,
  Salman Avestimehr, and Nageen Himayat. 2020.
\newblock Coded computing for low-latency federated learning over wireless edge
  networks.
\newblock \emph{IEEE Journal on Selected Areas in Communications}, (1).

\bibitem[{Raffel et~al.(2020)Raffel, Shazeer, Roberts, Lee, Narang, Matena,
  Zhou, Li, and Liu}]{t5}
Colin Raffel, Noam Shazeer, Adam Roberts, Katherine Lee, Sharan Narang, Michael
  Matena, Yanqi Zhou, Wei Li, and Peter~J Liu. 2020.
\newblock Exploring the limits of transfer learning with a unified text-to-text
  transformer.
\newblock \emph{Journal of Machine Learning Research}, (140).

\bibitem[{Rajpurkar et~al.(2016)Rajpurkar, Zhang, Lopyrev, and
  Liang}]{Rajpurkar2016SQuAD10}
Pranav Rajpurkar, Jian Zhang, Konstantin Lopyrev, and Percy Liang. 2016.
\newblock \href {https://doi.org/10.18653/v1/D16-1264} {{SQ}u{AD}: 100,000+
  questions for machine comprehension of text}.
\newblock In \emph{Proceedings of the 2016 Conference on Empirical Methods in
  Natural Language Processing}, pages 2383--2392, Austin, Texas. Association
  for Computational Linguistics.

\bibitem[{Ramaswamy et~al.(2019)Ramaswamy, Mathews, Rao, and
  Beaufays}]{Ramaswamy2019FederatedLF}
Swaroop~Indra Ramaswamy, Rajiv Mathews, K.~Rao, and Franccoise Beaufays. 2019.
\newblock Federated learning for emoji prediction in a mobile keyboard.
\newblock \emph{ArXiv}.

\bibitem[{Reddi et~al.(2021)Reddi, Charles, Zaheer, Garrett, Rush,
  Kone{\v{c}}n{\'y}, Kumar, and McMahan}]{reddi2020adaptive}
Sashank~J. Reddi, Zachary Charles, Manzil Zaheer, Zachary Garrett, Keith Rush,
  Jakub Kone{\v{c}}n{\'y}, Sanjiv Kumar, and Hugh~Brendan McMahan. 2021.
\newblock \href {https://openreview.net/forum?id=LkFG3lB13U5} {Adaptive
  federated optimization}.
\newblock In \emph{9th International Conference on Learning Representations,
  {ICLR} 2021, Virtual Event, Austria, May 3-7, 2021}. OpenReview.net.

\bibitem[{Regulation(2016)}]{gpdr}
General Data~Protection Regulation. 2016.
\newblock Regulation eu 2016/679 of the european parliament and of the council
  of 27 april 2016.
\newblock \emph{Official Journal of the European Union. Available at:
  http://ec. europa.
  eu/justice/data-protection/reform/files/regulation\_oj\_en. pdf (accessed 20
  September 2017)}.

\bibitem[{Reimers and Gurevych(2019)}]{Reimers2019SentenceBERTSE}
Nils Reimers and Iryna Gurevych. 2019.
\newblock \href {https://doi.org/10.18653/v1/D19-1410} {Sentence-{BERT}:
  Sentence embeddings using {S}iamese {BERT}-networks}.
\newblock In \emph{Proceedings of the 2019 Conference on Empirical Methods in
  Natural Language Processing and the 9th International Joint Conference on
  Natural Language Processing (EMNLP-IJCNLP)}, pages 3982--3992, Hong Kong,
  China. Association for Computational Linguistics.

\bibitem[{Ryffel et~al.(2018)Ryffel, Trask, Dahl, Wagner, Mancuso, Rueckert,
  and Passerat-Palmbach}]{ryffel2018generic}
Theo Ryffel, Andrew Trask, Morten Dahl, Bobby Wagner, Jason Mancuso, Daniel
  Rueckert, and Jonathan Passerat-Palmbach. 2018.
\newblock A generic framework for privacy preserving deep learning.
\newblock \emph{ArXiv preprint}.

\bibitem[{Sanh et~al.(2019)Sanh, Debut, Chaumond, and
  Wolf}]{Sanh2019DistilBERTAD}
Victor Sanh, Lysandre Debut, Julien Chaumond, and Thomas Wolf. 2019.
\newblock Distilbert, a distilled version of bert: smaller, faster, cheaper and
  lighter.
\newblock \emph{ArXiv}.

\bibitem[{Smith et~al.(2017)Smith, Chiang, Sanjabi, and
  Talwalkar}]{smith2017federated}
Virginia Smith, Chao{-}Kai Chiang, Maziar Sanjabi, and Ameet~S. Talwalkar.
  2017.
\newblock \href
  {https://proceedings.neurips.cc/paper/2017/hash/6211080fa89981f66b1a0c9d55c61d0f-Abstract.html}
  {Federated multi-task learning}.
\newblock In \emph{Advances in Neural Information Processing Systems 30: Annual
  Conference on Neural Information Processing Systems 2017, December 4-9, 2017,
  Long Beach, CA, {USA}}, pages 4424--4434.

\bibitem[{So et~al.(2020)So, G{\"u}ler, and Avestimehr}]{so2020byzantine}
Jinhyun So, Ba{\c{s}}ak G{\"u}ler, and A~Salman Avestimehr. 2020.
\newblock Byzantine-resilient secure federated learning.
\newblock \emph{IEEE Journal on Selected Areas in Communications}.

\bibitem[{So et~al.(2021{\natexlab{a}})So, G{\"u}ler, and
  Avestimehr}]{so2021codedprivateml}
Jinhyun So, Ba{\c{s}}ak G{\"u}ler, and A~Salman Avestimehr. 2021{\natexlab{a}}.
\newblock Codedprivateml: A fast and privacy-preserving framework for
  distributed machine learning.
\newblock \emph{IEEE Journal on Selected Areas in Information Theory}, (1).

\bibitem[{So et~al.(2021{\natexlab{b}})So, G{\"u}ler, and
  Avestimehr}]{so2021turbo}
Jinhyun So, Ba{\c{s}}ak G{\"u}ler, and A~Salman Avestimehr. 2021{\natexlab{b}}.
\newblock Turbo-aggregate: Breaking the quadratic aggregation barrier in secure
  federated learning.
\newblock \emph{IEEE Journal on Selected Areas in Information Theory}, (1).

\bibitem[{Stremmel and Singh(2020)}]{Stremmel2020PretrainingFT}
Joel Stremmel and Arjun Singh. 2020.
\newblock Pretraining federated text models for next word prediction.
\newblock \emph{ArXiv}.

\bibitem[{Sui et~al.(2020)Sui, Chen, Zhao, Jia, Xie, and Sun}]{Sui2020FedEDFL}
Dianbo Sui, Yubo Chen, Jun Zhao, Yantao Jia, Yuantao Xie, and Weijian Sun.
  2020.
\newblock \href {https://doi.org/10.18653/v1/2020.emnlp-main.165} {{F}ed{ED}:
  Federated learning via ensemble distillation for medical relation
  extraction}.
\newblock In \emph{Proceedings of the 2020 Conference on Empirical Methods in
  Natural Language Processing (EMNLP)}, pages 2118--2128, Online. Association
  for Computational Linguistics.

\bibitem[{Trischler et~al.(2017)Trischler, Wang, Yuan, Harris, Sordoni,
  Bachman, and Suleman}]{Trischler2017NewsQAAM}
Adam Trischler, Tong Wang, Xingdi Yuan, Justin Harris, Alessandro Sordoni,
  Philip Bachman, and Kaheer Suleman. 2017.
\newblock \href {https://doi.org/10.18653/v1/W17-2623} {{N}ews{QA}: A machine
  comprehension dataset}.
\newblock In \emph{Proceedings of the 2nd Workshop on Representation Learning
  for {NLP}}, pages 191--200, Vancouver, Canada. Association for Computational
  Linguistics.

\bibitem[{Vaswani et~al.(2017)Vaswani, Shazeer, Parmar, Uszkoreit, Jones,
  Gomez, Kaiser, and Polosukhin}]{vaswani2017attentionia}
Ashish Vaswani, Noam Shazeer, Niki Parmar, Jakob Uszkoreit, Llion Jones,
  Aidan~N. Gomez, Lukasz Kaiser, and Illia Polosukhin. 2017.
\newblock \href
  {https://proceedings.neurips.cc/paper/2017/hash/3f5ee243547dee91fbd053c1c4a845aa-Abstract.html}
  {Attention is all you need}.
\newblock In \emph{Advances in Neural Information Processing Systems 30: Annual
  Conference on Neural Information Processing Systems 2017, December 4-9, 2017,
  Long Beach, CA, {USA}}, pages 5998--6008.

\bibitem[{Wang et~al.(2020{\natexlab{a}})Wang, Sreenivasan, Rajput,
  Vishwakarma, Agarwal, Sohn, Lee, and Papailiopoulos}]{wang2020attack}
Hongyi Wang, Kartik Sreenivasan, Shashank Rajput, Harit Vishwakarma, Saurabh
  Agarwal, Jy{-}yong Sohn, Kangwook Lee, and Dimitris~S. Papailiopoulos.
  2020{\natexlab{a}}.
\newblock \href
  {https://proceedings.neurips.cc/paper/2020/hash/b8ffa41d4e492f0fad2f13e29e1762eb-Abstract.html}
  {Attack of the tails: Yes, you really can backdoor federated learning}.
\newblock In \emph{Advances in Neural Information Processing Systems 33: Annual
  Conference on Neural Information Processing Systems 2020, NeurIPS 2020,
  December 6-12, 2020, virtual}.

\bibitem[{Wang et~al.(2020{\natexlab{b}})Wang, Yurochkin, Sun, Papailiopoulos,
  and Khazaeni}]{wang2020federated}
Hongyi Wang, Mikhail Yurochkin, Yuekai Sun, Dimitris~S. Papailiopoulos, and
  Yasaman Khazaeni. 2020{\natexlab{b}}.
\newblock \href {https://openreview.net/forum?id=BkluqlSFDS} {Federated
  learning with matched averaging}.
\newblock In \emph{8th International Conference on Learning Representations,
  {ICLR} 2020, Addis Ababa, Ethiopia, April 26-30, 2020}. OpenReview.net.

\bibitem[{Wolf et~al.(2020)Wolf, Debut, Sanh, Chaumond, Delangue, Moi, Cistac,
  Rault, Louf, Funtowicz, Davison, Shleifer, von Platen, Ma, Jernite, Plu, Xu,
  Le~Scao, Gugger, Drame, Lhoest, and Rush}]{Wolf2020TransformersSN}
Thomas Wolf, Lysandre Debut, Victor Sanh, Julien Chaumond, Clement Delangue,
  Anthony Moi, Pierric Cistac, Tim Rault, Remi Louf, Morgan Funtowicz, Joe
  Davison, Sam Shleifer, Patrick von Platen, Clara Ma, Yacine Jernite, Julien
  Plu, Canwen Xu, Teven Le~Scao, Sylvain Gugger, Mariama Drame, Quentin Lhoest,
  and Alexander Rush. 2020.
\newblock \href {https://doi.org/10.18653/v1/2020.emnlp-demos.6} {Transformers:
  State-of-the-art natural language processing}.
\newblock In \emph{Proceedings of the 2020 Conference on Empirical Methods in
  Natural Language Processing: System Demonstrations}, pages 38--45, Online.
  Association for Computational Linguistics.

\bibitem[{Yang et~al.(2019)Yang, Liu, Chen, and Tong}]{yang2019federated}
Qiang Yang, Yang Liu, Tianjian Chen, and Yongxin Tong. 2019.
\newblock Federated machine learning: Concept and applications.
\newblock \emph{ACM Transactions on Intelligent Systems and Technology (TIST)},
  (2).

\bibitem[{Yang et~al.(2018{\natexlab{a}})Yang, Andrew, Eichner, Sun, Li, Kong,
  Ramage, and Beaufays}]{Yang2018APPLIEDFL}
T.~Yang, G.~Andrew, Hubert Eichner, Haicheng Sun, W.~Li, Nicholas Kong,
  D.~Ramage, and F.~Beaufays. 2018{\natexlab{a}}.
\newblock Applied federated learning: Improving google keyboard query
  suggestions.
\newblock \emph{ArXiv}.

\bibitem[{Yang et~al.(2018{\natexlab{b}})Yang, Qi, Zhang, Bengio, Cohen,
  Salakhutdinov, and Manning}]{Yang2018HotpotQAAD}
Zhilin Yang, Peng Qi, Saizheng Zhang, Yoshua Bengio, William Cohen, Ruslan
  Salakhutdinov, and Christopher~D. Manning. 2018{\natexlab{b}}.
\newblock \href {https://doi.org/10.18653/v1/D18-1259} {{H}otpot{QA}: A dataset
  for diverse, explainable multi-hop question answering}.
\newblock In \emph{Proceedings of the 2018 Conference on Empirical Methods in
  Natural Language Processing}, pages 2369--2380, Brussels, Belgium.
  Association for Computational Linguistics.

\bibitem[{Zhu et~al.(2019)Zhu, Liu, and Han}]{zhu2020deep}
Ligeng Zhu, Zhijian Liu, and Song Han. 2019.
\newblock \href
  {https://proceedings.neurips.cc/paper/2019/hash/60a6c4002cc7b29142def8871531281a-Abstract.html}
  {Deep leakage from gradients}.
\newblock In \emph{Advances in Neural Information Processing Systems 32: Annual
  Conference on Neural Information Processing Systems 2019, NeurIPS 2019,
  December 8-14, 2019, Vancouver, BC, Canada}, pages 14747--14756.

\bibitem[{Zhu et~al.(2020)Zhu, Wang, Hong, and Xiao}]{Zhu2020EmpiricalSO}
Xinghua Zhu, Jianzong Wang, Zhenhou Hong, and Jing Xiao. 2020.
\newblock \href {https://doi.org/10.18653/v1/2020.findings-emnlp.55} {Empirical
  studies of institutional federated learning for natural language processing}.
\newblock In \emph{Findings of the Association for Computational Linguistics:
  EMNLP 2020}, pages 625--634, Online. Association for Computational
  Linguistics.

\end{thebibliography}
\bibliographystyle{acl_natbib}


\clearpage
\appendix

\noindent
{\Large{\textit{\textbf{Appendix}}}} \\
\smallskip

\section{FL+NLP}
\label{ssec:motivation}

Many realistic NLP services heavily rely on users' local data (e.g., text messages, documents and their tags, questions and selected answers, etc.), which can be located at either personal devices or larger data-silos for organizations.
These local data are usually regarded as highly private and thus not directly accessible by anyone, according to many data privacy regulations; this makes it difficult to train a high-performance model to benefit users.
Federated learning aims to solve machine learning under such a privacy-preserving use case, thus offering a novel and promising direction to the community: FL+NLP.

Apart from the goal of learning a shared global model for all clients, FL also provides a new perspective for many other interesting research questions in NLP.
One related direction is to develop personalized models for NLP applications, which requires both protection of data privacy and transferred ability on users' own input feature distribution caused by language styles, interested topics and so on.
The recent concerns on adversarial attacks and safety issues of NLP models are also highly related to FL+NLP.
We thus believe FL+NLP is of vital importance for applying NLP technologies in realistic use cases and could benefit many relevant research areas.

\subsection{Challenges of Applying FL in NLP}
\label{ssec:challenge}

Given the promising benefits of studying FL+NLP, however, this research direction is currently blocked by the lack of a standardized platform providing fundamental building blocks:   benchmark datasets, NLP models, FL methods, evaluation protocols, etc.
Most of the current FL platforms either focus on unifying various FL methods and use computer vision models and datasets for their experiments, but lack  the ability to connect the study of pre-trained language models, the most popular NLP , and realistic NLP applications of various task formulations.

The first challenge in developing a comprehensive and universal platform for FL+NLP is to deal with various task formulations for realistic NLP applications, which have different input and output formats (Section~\ref{ssec:nlptasks}).
As the non-IID data partition over clients is the major feature of FL problems, it is also a challenge to simulate the realistic non-IID partition for existing NLP datasets (Section~\ref{ssec:partition}).
Finally, a platform also must integrate various FL methods with the Transformer-based NLP models for a variety of task types, and thus a flexible and extensible learning framework is needed. 
In particular, the conventional trainer component of Transformers now needs to be modified for efficient and safe communications towards federated learning (Section~\ref{sec:framework}).

\section{Basic Formulations of NLP Tasks}
\label{ssec:nlptasks}
There are various types of NLP applications, but many of them share a similar task formulation (i.e., input-and-put formats). 
We show four common task formulations that can cover most of the mainstream NLP applications: text classification, sequence tagging, question answering, sequence-to-sequence generation.

\smallskip \noindent \textbf{Text Classification (TC)}
The input is a sequence of words,  $x=[w_1, w_2, \dots]$, and the output is a label $y$ in a fixed set of labels $\mathcal{L}$. 
Many NLP applications can be formulated as text classification tasks. 
For example, we can use TC models for classifying the topic of a news article to be \textit{political}, \textit{sports}, \textit{entertainment}, etc., or analyzing movie reviews to be \textit{positive}, \textit{negative} or \textit{neutral}. 

\smallskip \noindent \textbf{Sequence Tagging (ST)}
The input is a sequence of words,  $x=[w_1, w_2, \dots, w_N]$, and the output is a same-length sequence of tags $y=[t_1, t_2, \dots, t_N]$, where $t_i$ is in a fixed set of labels $\mathcal{L}$.
The main difference between TC and ST is that ST learns to classify the label of each token in a sentence, which is particularly useful in analyzing syntactic structures (e.g., part-of-speech analysis, phrase chunking, and word segmentation) and extracting spans (e.g., named entity recognition).

\smallskip \noindent \textbf{Question Answering (QA)}
Given a passage $P=[w_1, w_2, \dots, w_N]$ and a question $q$ as input, the task is to locate a span in the passage as the answer to the question.
Thus, the output is a pair of token index $(s, e)$ where $s, e \in \{1, 2, \dots, N\}$ for denoting the begin and end of the span in the passage.
This particular formulation is also known as \textit{reading comprehension}.

\smallskip \noindent \textbf{Natural Language Generation (NLG)}
Both input and output are sequence of words, $x=[w_1^i, w_2^i, \dots, w_N^i]$ , $y=[w_1^o, w_2^o, \dots, w_M^o]$. 
It is shared by many realistic applications such as summarization, response generation in dialogue systems, machine translation, etc. 

\smallskip \noindent \textbf{Language Modeling (LM)} 
The left-to-right language modeling task considers a sequence of words as the input $x=[w_1, w_2, \dots, w_n]$ and a token $y=w_{n+1}$ as the output. 
The output token is expected to be the most plausible next word of the incomplete sentence denoted as $x$.
Although the direct application of LM is limited, 
a high-performance pre-trained language model can benefit a wide range of NLP applications (as above) via fine-tuning.
It  also serves as an excellent test bed as it requires no human annotations at all.

\begin{figure}[t]
\centering
{\includegraphics[width=0.85\linewidth]{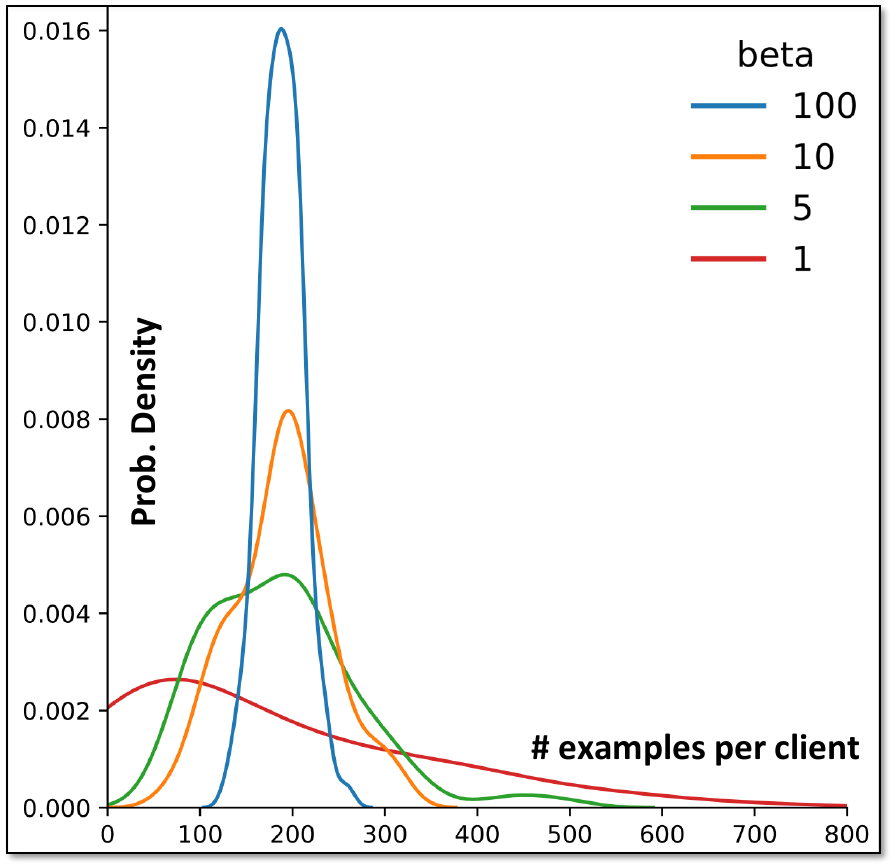}}
\caption{The probability density of quantity of training examples in each of the 100 clients on the \textit{20News} dataset with different $\beta$. When $\beta$ is larger, then all clients share more similar numbers of examples; when $\beta$ is smaller, then the range of the quantity is much wider --- i.e., the larger differences between clients in terms of their sizes of datasets.}
\label{fig:quantityshift}
\end{figure}

\smallskip
\noindent
\textbf{Others.} There are some other applications that not are covered by the above four basic formulations, and our extensible platform (detailed in Section~\ref{sec:framework}) enables users to easily implement their specific tasks. 
For each task formulation, we show which datasets are used in FedNLP and how we partition them in Section~\ref{sec:data}.


\section{Implementation Details}
\paragraph{Non-IID. Label Distribution}
Note that this might cause a few clients not to have enough examples to sample for particular labels if they are already used up. 
Prior works choose to stop assigning early and remove such clients, but it consequently loses the other unused examples and also causes the inconsistency of client numbers.
Thus, to avoid these issues, we propose a \textit{dynamic reassigning} method which complement the vacancy of a label by filling in the examples of other labels based on their current ratio of remaining unassigned examples.

\subsection{The FedNLP Training Pipeline: Security and Efficiency}
\label{ssec:details}
Under the  definition of federated learning in \Cref{algo:generalized_fedavg}, we design a training system to support the research of NLP in the FL paradigm. We highlight its core capabilities and design as follows.

\paragraph{Supporting diverse FL algorithms.} \texttt{FedNLP} aims to enable flexible customization for future algorithmic innovations. We have supported a number of classical federated learning algorithms, including FedAvg \cite{mcmahan2017communication}, FedOPT \cite{reddi2020adaptive}, and FedProx \cite{li2018federated}. These algorithms follow the same framework introduced in Algorithm \ref{algo:generalized_fedavg}. The algorithmic APIs are modularized: all data loaders follow the same format of input and output arguments, which are compatible with different models and algorithms and are easy to support new datasets; the method of defining the model and related trainer is kept the same as in centralized training to reduce the difficulty of developing the distributed training framework. For new FL algorithm development, worker-oriented programming reduces the difficulty of message passing and definition. More details are introduced in Appendix \ref{app:algorithm}. 

\paragraph{Enabling secure benchmarking with lightweight secure aggregation.} In particular, \texttt{FedNLP} enhances the security aspect of federated training, which is not supported by existing non-NLP-oriented benchmarking libraries (e.g., TFF, LEAF). This is motivated by the fact that model weights from clients may still have the risk of privacy leakage \cite{zhu2020deep}. To break this barrier, we integrate secure aggregation (SA) algorithms to the \texttt{FedNLP} system. NLP researchers do not need to master security-related knowledge and also benefit from a secure distributed training environment. To be more specific, \texttt{FedNLP} supports state-of-the-art SA algorithms \texttt{LightSecAgg}, \texttt{SecAgg} \cite{bonawitz2017practical}, and \texttt{SecAgg}+ \cite{bell2020secure}. At a  high-level understanding, SA protects the client model by generating a single random mask and allows their cancellation when aggregated at the server. Consequently, the server can only see the aggregated model and not the raw model from each client. In this work, our main effort is to design and optimize these SA algorithms in the context of the \texttt{FedNLP} system. We provide an algorithmic performance comparison in Appendix \ref{app:lightsecagg}. 

\paragraph{Realistic evaluation with efficient distributed system design.} \texttt{FedNLP} aims to support distributed training in multiple edge servers (e.g, AWS EC2) or edge devices (e.g., IoTs and smartphones). To achieve this, the system is designed with three layers: the application layer, the algorithm layer, and the infrastructure layer.
At the application layer, FedNLP provides three modules: 
data management, model definition, and a single-process trainer for all task formats;
at the algorithm layer, FedNLP supports various FL algorithms; 
at the infrastructure layer, 
FedNLP aims at integrating single-process trainers with a distributed learning system for FL. 
Specifically, we make each layer and module perform its own duties and have a high degree of modularization. We refer readers to Appendix \ref{sec:framework} for a detailed description of the system architecture and design philosophy.

\section{More Related Works}

\paragraph{Federated Learning Methods.} Federated Learning (FL) is a widely disciplinary research area that mainly focuses on three aspects: statistical challenge, trustworthiness, and system optimization. Numerous methods have been proposed to solve statistical challenges, including FedAvg \cite{McMahan2017CommunicationEfficientLO}, FedProx \cite{li2018federated}, FedOPT \cite{reddi2020adaptive}, FedNAS \cite{he2020fednas,MiLeNAS2020}, and FedMA \cite{wang2020federated} that alleviate the non-IID issue with distributed optimization, and new formulations, MOCHA \cite{smith2017federated}, pFedMe \cite{dinh2020personalized}, perFedAvg \cite{fallah2020personalized}, and Ditto \cite{Li2020DittoFA}, that consider personalization and fairness in federated training. 

For trustworthiness, security and privacy are the two main research directions that are mainly concerned with resisting data or model attacks, reconstruction, and leakage during training \cite{so2021turbo,so2021codedprivateml,so2020byzantine,prakash2020coded,prakash2020mitigating,Elkordy2020SecureAW,prakash2020coded,wang2020attack,lyu2020privacy}. 
Given that modern deep neural networks are over-parameterized and dominate nearly all learning tasks, researchers also proposed algorithms or systems to improve the efficiency and scalability of edge training \cite{he2020group,he2020fedml,he2019central,He2021FedGraphNNAF}. We refer readers to the canonical survey \cite{kairouz2019advances} for details.

Although tremendous progress has been made in the past few years, these algorithms or systems have not been fully evaluated on realistic NLP tasks introduced in this paper. 

\section{Future Directions}
\label{app:future_directions}


 
    \paragraph{Minimizing the performance gap.} In the FL setting, we demonstrate that federated fine-tuning still has a large accuracy gap in the non-IID dataset compared to centralized fine-tuning. Developing algorithms for Transformer models with NLP tasks is of the highest priority.

    \paragraph{Improving the system efficiency and scalability.} 
    Transformer models are usually large, while resource-constrained edge devices may not be able to run large models. 
    Designing efficient FL methods for NLP tasks is thus a practical problem worth solving. 
    How to adopt a reasonable user selection mechanism to avoid stragglers and speed up the convergence of training algorithms is also a pressing problem to be solved.
    
    \paragraph{Trustworthy and privacy-preserving NLP.} 
    We argue that it is an important future research direction to analyze and assure the privacy-preserving ability of these methods, although our focus in this paper is the implementation and performance analysis of the FL methods for NLP tasks.
    It is now an open problem for both FL and NLP areas, while it is  an orthogonal goal for improving the trustworthy of decentralized learning, and it is only possible to study privacy preservation when we have an existing FL+NLP platform. This is also part of our motivation in proposing FedNLP, and we believe our framework provides a set of flexible interfaces for future development to analyze and improve the privacy-preserving ability of FL methods for NLP tasks and beyond.

      \paragraph{Personalized FedNLP.} From the perspective of the data itself, user-generated text is inherently personalized. Designing personalized algorithms to improve model accuracy or fairness is a very promising direction. In addition, it is also an interesting problem to adapt the heterogeneous model architecture for each client in the FL network. We show that it is feasible to only fine-tune a small amount of the parameters of LMs, so it is promising to adapt recent prefix-tuning methods~\cite{Li2021PrefixTuningOC} for personalizing the parameters of NLP models within the FedNLP framework.
    


\section{The System Design of FedNLP}
\label{sec:framework}




\begin{figure*}[t]
\centering
{\includegraphics[width=1\linewidth]{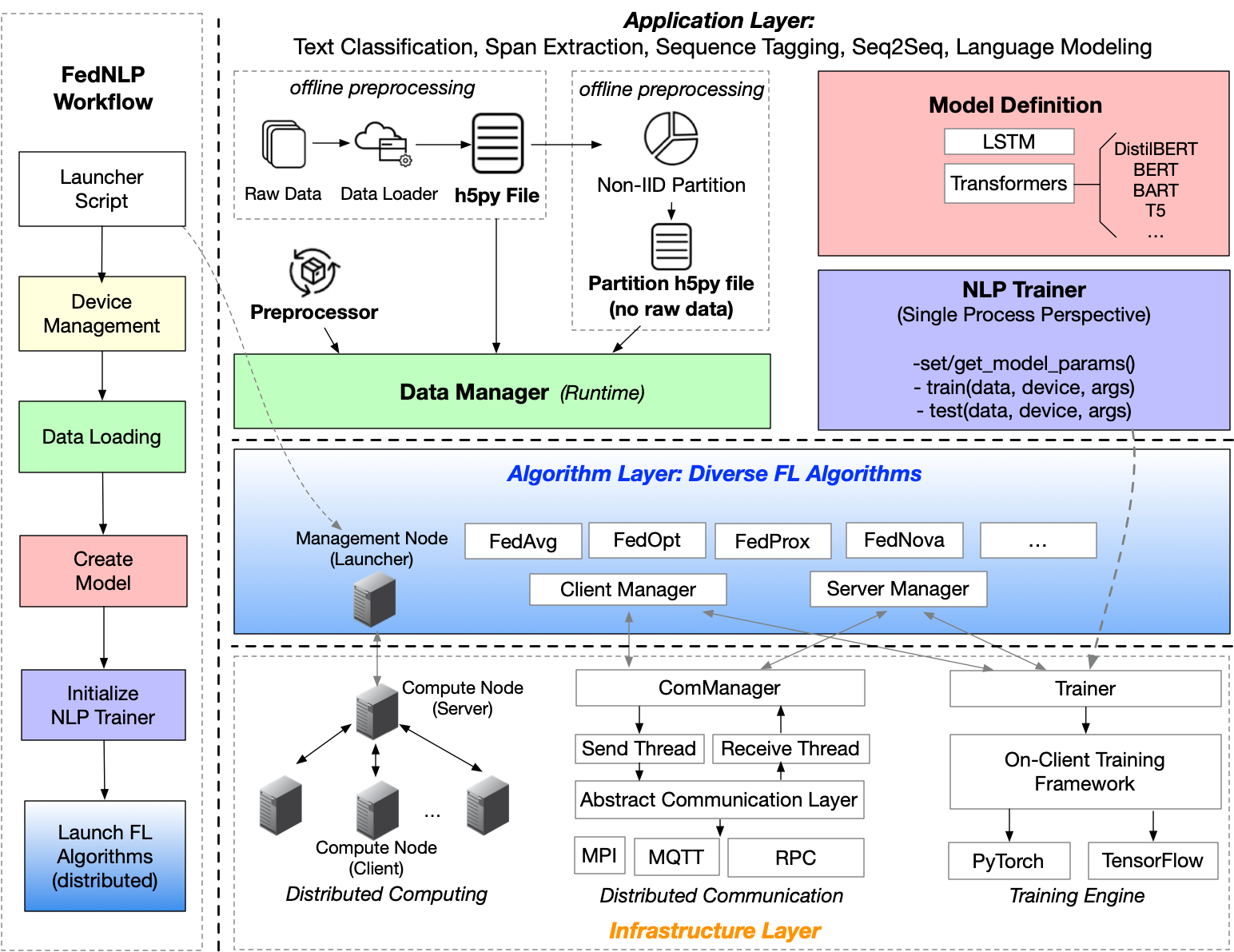}}
\caption{The overall workflow and system design of the proposed {FedNLP} platform. }
\label{fig:fednlp_framework}
\end{figure*}

The FedNLP platform consists of three layers: 
the application layer, the algorithm layer, and the infrastructure layer. 
At the application layer, FedNLP provides three modules: 
data management, model definition, and single-process trainer for all task formats; 
At the algorithm layer, FedNLP supports various FL algorithms; 
At the infrastructure layer, 
FedNLP aims at integrating single-process trainers with a distributed learning system for FL. 
Specifically, we make each layer and module perform its own duties and have a high degree of modularization. 

\subsection{Overall Workflow}
The module calling logic flow of the whole framework is shown on the left of Figure \ref{fig:fednlp_framework}. 
When we start the federated training, 
we first complete the launcher script, device allocation, data loading, and model creation, and finally call the API of the federated learning algorithm. This process is expressed in Python-style code (see Alg.~\ref{alg:code}).

\begin{algorithm}[t]
\caption{The FedNLP Workflow}
\label{alg:code}
\definecolor{codeblue}{rgb}{0.25,0.5,0.5}
\definecolor{codekw}{rgb}{0.85, 0.18, 0.50}
\lstset{
  backgroundcolor=\color{white},
  basicstyle=\fontsize{7.5pt}{7.5pt}\ttfamily\selectfont,
  columns=fullflexible,
  breaklines=true,
  captionpos=b,
  commentstyle=\fontsize{7.5pt}{7.5pt}\color{codeblue},
  keywordstyle=\fontsize{7.5pt}{7.5pt}\color{codekw},
  keywords={map_process_to_gpu},
  keywords=[2]{TCDataManager},
  keywords=[3]{TCDataManager},
  keywordstyle={\color{blue!80!black}},
  keywordstyle=[2]{\color{red!80!green}},
  keywordstyle=[3]{\color{green!50!black}},
}
\begin{lstlisting}[language=python]
# using text classification (TC) as an example

# initialize distributed computing environment
process_id, ... = FedNLP_init()

# GPU device management
device =  map_process_to_gpu(process_id, ...)

# data management 
data_manager = TCDataManager (process_id, ...) 
# load the data dictionary by process_id
data_dict = dm.load_federated_data(process_id)

# create model by specifying the task
client_model, ... = create_model(model_args,                                                        formulation="classification")

# define a customized NLP Trainer
client_trainer = TCTrainer(device, 
                           client_model, ...)

# launch the federated training (e.g., FedAvg)
FedAvg_distributed(..., device, 
                        client_model, 
                        data_dict, ..., 
                        client_trainer) 

\end{lstlisting}
\end{algorithm}

\subsection{The Application Layer}
\paragraph{Data Management.} In data management, What \code{DataManager} does is control the whole workflow from loading data to returning trainable features. To be specific, \code{DataManager} is set up for reading h5py data files and driving a \code{preprocessor} to convert raw data to features. There are four types of \code{DataManager} according to the task definition. Users can customize their \code{DataManager} by inheriting one of the \code{DataManager} class, specifying data operation functions, and embedding a particular preprocessor. Note that the raw data's \code{H5Py} file and the non-IID partition file are preprocessed offline, while \code{DataManager} only loads them in runtime.

\paragraph{Model Definition.} We support two types of models: Transformer and LSTM. For Transformer models, to dock with the existing NLP ecology, our framework is compatible with the \textit{HuggingFace Transformers}  library~\cite{Wolf2020TransformersSN}, 
so that various types of Transformers can be directly reused without the need for re-implementation. 
Specifically, our code is compatible with the three main classes of \code{Tokenizer}, \code{Model}, and \code{Config} in \textit{HuggingFace}. 
Users can also customize them based on \code{HuggingFace}'s code. Although LSTM has gradually deviated from the mainstream, we still support LSTM to reflect the framework's integrity, which may meet some particular use cases in a federated setting.

\paragraph{NLP Trainer (single process perspective).} As for the task-specific NLP \code{Trainer}, 
the most prominent feature is that it does not require users to have any background in distributed computing. 
Users of FedNLP only need to complete single-process code writing. 
A user should inherit the \code{Trainer} class in the application layer to implement the four methods as shown in the figure: 1. the \code{get\_model\_params()} interface allows the algorithm layer to obtain model parameters and transmit them to the server; 2. the \code{set\_model\_params()} interface obtains the updated model from the server's aggregation and then updates the model parameters of the local model; 3. the programming of the \code{train()} and \code{test()} function only needs to consider the data of a single user, meaning that the trainer is completely consistent with the centralized training.

\subsection{The Algorithm Layer}
\label{app:algorithm}

In the design of the algorithm layer, we follow the principle of one-line API. The parameters of the API include model, data, and single-process trainer (as shown in Algorithm~\ref{alg:code}). The algorithms we support include:

\paragraph{{Centralized Training.}} We concatenate all client datasets and use the global data $\mathcal{D}_G$ to train a global model --- i.e., the conventional protocol for learning an NLP model on a dataset.

\paragraph{FedAvg}\cite{mcmahan2017communication} is the \textit{de facto} method for federated learning, assuming both client and server use the \textit{SGD} optimizer for updating model weights.

\paragraph{FedProx}\cite{li2018federated} can tackle statistical heterogeneity by restricting the local model updates to be closer to the initial (global) model with L2 regularization for better stability in training.
    
\paragraph{FedOPT}\cite{reddi2020adaptive} is a generalized version of \texttt{FedAvg}. 
There are two gradient-based optimizers in the algorithm: 
\texttt{ClientOpt} and \texttt{ServerOpt} (please refer to the pseudo code in the original paper \cite{reddi2020adaptive}). 
While \texttt{ClientOpt} is used to update the local models, \texttt{SerevrOpt} treats the negative of aggregated local changes $-\Delta^{(t)}$ as a pseudo-gradient and applies it on the global model. 
In our FedNLP framework, by default, 
we set the \texttt{ClientOpt} to be AdamW~\cite{loshchilov2017decoupled} and the \texttt{SerevrOpt} to be SGD with momentum (0.9) and fix server learning rate as $1.0$.

Each algorithm includes two core objects, \code{ServerManager} and \code{ClientManager}, which integrate the communication module \code{ComManager} from the infrastructure layer and the \code{Trainer} of the training engine to complete the distributed algorithm protocol and edge training. Note that users can customize the Trainer by passing a customized \code{Trainer} through the algorithm API.

\subsection{The Infrastructure Layer}
The infrastructure layer includes three modules:

\smallskip
\noindent
1) Users can write distributed scripts to manage GPU resource allocation. 
In particular, FedNLP provides the GPU assignment API (\code{map\_process\_to\_gpu()} in Algorithm \ref{alg:code}) to assign specific GPUs to different FL Clients. 

\smallskip
\noindent
2) The algorithm layer can use a unified and abstract \texttt{ComManager} to complete a complex algorithmic communication protocol. 
Currently, we support MPI (Message Passing Interface), RPC (Remote procedure call), and MQTT (Message Queuing Telemetry Transport) communication backend. MPI meets the distributed training needs in a single cluster; RPC meets the communication needs of cross-data centers (e.g., cross-silo federated learning); MQTT can meet the communication needs of smartphones or IoT devices.

\smallskip
\noindent
3) The third part is the training engine, which reuses the existing deep learning training engines by presenting as the \code{Trainer} class. Our current version of this module is built on \code{PyTorch}, but it can easily support frameworks such as \code{TensorFlow}. 
In the future, we may consider supporting the lightweight edge training engine optimized by the compiler technology at this level.

\subsection{Enhancing Security with Secure Aggregation (SA)}
\label{app:lightsecagg}

\texttt{FedNLP} supports state-of-the-art SA algorithms \texttt{LightSecAgg}, \texttt{SecAgg} \cite{bonawitz2017practical}, and \texttt{SecAgg+} \cite{bell2020secure}. Here, we provide a short performance comparison of these three algorithms. In general, \texttt{LightSecAgg} provides the same model privacy guarantees as SecAgg \cite{bonawitz2017practical} and SecAgg+ \cite{bell2020secure}) while substantially reducing the aggregation (hence run-time) complexity (Figure \ref{fig:runtime_CNN_varDropout}). The main idea of \texttt{LightSecAgg} are that each user protects its local model using a locally generated random mask. This mask is then encoded and shared with other users, in such a way that the aggregate mask of any sufficiently large set of surviving users can be directly reconstructed at the server. Our main effort in \texttt{FedNLP} is integrating  these algorithms, optimizing its system performance, and designing user-friendly APIs to make them compatible with NLP models and FL algorithms.


\end{document}